\documentclass[11pt]{article}
\usepackage[margin=1in]{geometry}
\usepackage[T1]{fontenc}
\usepackage{lmodern}
\AtBeginDocument{%
  \DeclareFontShape{T1}{lmr}{bx}{sc}{<->ssub*cmr/bx/sc}{}%
  \DeclareFontShape{T1}{lmr}{m}{scit}{<->ssub*lmr/m/sc}{}}
\usepackage{microtype}
\usepackage{amsmath,amssymb,amsthm}
\usepackage{booktabs}
\usepackage{graphicx}
\usepackage{xcolor}
\usepackage{colortbl}
\definecolor{nmtint}{HTML}{007D55}
\AddToHook{env/table/begin}{\setlength{\belowcaptionskip}{5pt}}
\newcommand{\stackhead}[2]{\begin{tabular}[b]{@{}c@{}}#1\\#2\end{tabular}}
\usepackage{algorithm}
\usepackage{algpseudocode}
\usepackage[numbers,sort&compress]{natbib}
\usepackage[colorlinks=true,linkcolor=blue!50!black,citecolor=blue!50!black,urlcolor=blue!50!black]{hyperref}
\usepackage[capitalise,noabbrev]{cleveref}
\usepackage{enumitem}
\usepackage{float}
\setlist{itemsep=2pt,topsep=3pt}
\definecolor{newpart}{HTML}{007D55}
\newcommand{\newpart}[1]{\textcolor{newpart}{#1}}
\usepackage[xcolor,leftbars]{changebar}
\makeatletter\let\@footnotetext\ltx@footnotetext\let\@mpfootnotetext\ltx@mpfootnotetext\makeatother
\cbcolor{newpart}
\newtheorem{proposition}{Proposition}
\theoremstyle{remark}

\newcommand{\R}{\mathbb{R}}
\newcommand{\E}{\mathbb{E}}
\newcommand{\nearest}{\operatorname{nearest}}
\newcommand{\Lib}{\textsc{count}}
\newcommand{\Dist}{\textsc{distance}}
\newcommand{\Cost}{\textsc{cost}}
\newcommand{\Hyb}{\textsc{cost+distance}}
\newcommand{\Mean}{\textsc{net-mean}}
\newcommand{\Ours}{\textsc{net-median}}
\newcommand{\OursFH}{\textsc{net-median-fh}}
\DeclareMathOperator*{\argmin}{arg\,min}

\title{Revive What Pays: Evidence-Weighted Net-Gain Revival\\ for Mini-Batch k-Means}
\author{Eyal Weiss\\ Robotics Institute, Carnegie Mellon University}
\date{}

\begin{document}
\maketitle

\begin{abstract}
Mini-batch k-means algorithms such as e-means periodically move rarely used centers to where they reduce the error
more, a step called revival. The strongest rule so far, the cost-based revival proposed for the FAISS library, compares
running estimates of the cost of removing a center and of the gain of splitting another. These estimates are precise
for large, long-lived clusters and mostly noise for small or young ones, and the rule trusts them alike. We treat
each move as a decision under uncertainty: a center is moved only if the predicted gain of the split it funds is more
likely than not to exceed the predicted cost of its removal. Each estimate is shrunk toward the population of centers
according to its evidence (Gamma--Poisson rates, empirical-Bayes cost ratios, and log-gains with a precision-weighted
Paule--Mandel prior), and the decision uses posterior medians, which, unlike posterior means, give no credit to centers
about which nothing is known. The rule has no tuning parameter. In a pre-registered study on 14 datasets, 34
dataset--size combinations and 20 previously unused seeds, it lowers the test error by 2.9\% and the training error by
8.9\% relative to cost-based revival, and by 5\% and 7\% with five epochs, where an earlier version based on posterior
means broke down. With its decision in C++ it takes about 7\% longer than cost-based revival; an automatic early end of
the move phase, which stops revival once it has stopped paying, saves about 5\% of the running time with five or ten
epochs at a cost of at most 0.33\% in error. The rule is a function of per-center statistics only and
runs unchanged inside GPU e-means and o-means.
\end{abstract}

\section{Introduction}\label{sec:intro}

The k-means problem asks for $k$ centers that minimize the mean squared distance from the points of a dataset to their
nearest center. Mini-batch algorithms solve it on large datasets by updating the centers from small random batches of
points~\citep{sculley2010}. A recent algorithm of this kind is e-means~\citep{bessa2026emeans}, now part of the FAISS
library~\citep{douze2024faiss}. In e-means, each center is an exponential moving average (EMA) of the points assigned to
it, with a step size and a batch size set by scaling laws. The algorithm reaches the error of Lloyd's algorithm in about a
tenth of the passes over the data.

\begin{figure}[t]
\centering
\includegraphics{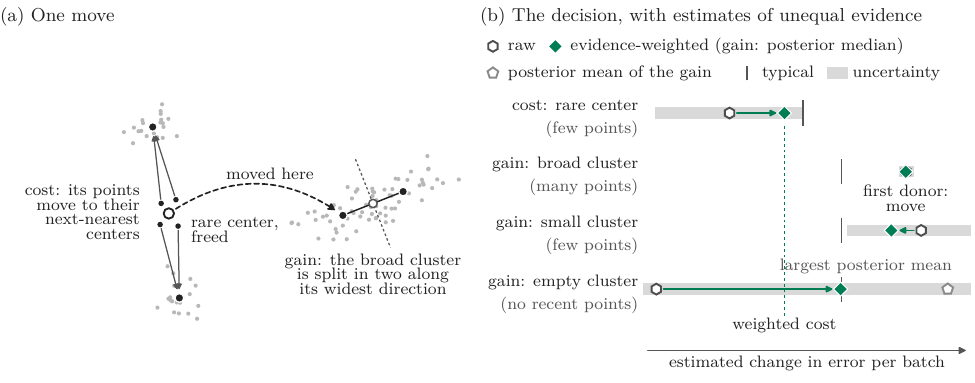}
\caption{One revival move (a schematic: synthetic points, illustrative positions, no results). (a)~Freeing a rare
center moves each of its points to its next-nearest center, which raises the error (the \emph{removal cost} $C_j$;
\cref{sec:method}). The freed center splits a broad cluster along its direction of largest spread, which lowers the
error (the \emph{split gain} $G_s$). (b)~An estimate from few points is uncertain (gray bars). \Ours{} shrinks each
estimate toward the population of centers according to its evidence (green; \cref{sec:estimation}) and makes the move
only if the estimated gain exceeds the estimated cost, $\tilde C_j<\tilde G_s$. For the gain, it takes the posterior
median. Here, the raw gains would favor a small cluster with little evidence, and the posterior mean (pentagon;
our first version, \cref{sec:why}) a cluster without recent points, about which nothing is known; the median favors
the broad cluster, which becomes the first donor.}\label{fig:teaser}
\end{figure}

Like most online k-means methods, e-means also moves centers during training. Every five batches, it frees the centers
of clusters that have become rare and places each of them next to the center of a large cluster, splitting that
cluster in two. Its authors call this step \emph{dead-cluster revival}. Revival matters because a center in a sparse
region contributes little, while a large cluster with a wide spread contributes much of the error. Thus, moving the
first to help the second lowers the error. This paper is about the following question: Which centers should a revival
step free, and which clusters should it split?

\paragraph{Three Rules So Far.} e-means frees a center when its running count of points falls below a tenth of the
average (\Lib). Counting alone also frees small clusters that are far from every other center, and their points are
then far from their next center. Our earlier rule keeps such clusters by adding a distance test
(\Dist;~\citep{weiss2026small}). On the same day as that work, a pull request to FAISS, since merged, proposed a rule that estimates
the two quantities a move trades off~\citep{faiss2026pr5915}. The \emph{removal cost} of a center is the increase in
error when its points move to their second-nearest center, and the \emph{split gain} of a cluster is the decrease in
error when its center is split along the direction of largest spread. The rule frees centers whose removal cost is
small relative to the median cost and splits the clusters with the largest gains (\Cost). On our benchmarks, \Cost{} has
much lower errors than both earlier rules (\cref{sec:experiments}).

\paragraph{Unequal Evidence.} Both quantities are estimated through running averages over the points that a center has
received. For a large cluster that has existed for many batches, the estimates are precise. For a center that receives
one point every few batches, or that was created by a split a few batches ago, they are mostly noise. However, a rule
that compares raw estimates treats both kinds alike. Because it selects the extremes (the cheapest centers to free and
the largest gains), it selects the noisiest estimates preferentially and acts on their errors in the favorable
direction, a form of the optimizer's curse~\citep{smith2006optimizer}. Indeed, in a calibration against the exact values on
the full data (\cref{app:calibration}), the removal costs of low-evidence centers were underestimated by one to two
orders of magnitude, and the gains of the chosen donors were overestimated by a factor of 1.2 to~2.1.

\paragraph{Contributions.} This paper makes four contributions. (i)~We make revival a decision under uncertainty and
give a rule, \Ours, that weighs every estimate by its evidence and has no tuning parameter beyond numerical floors.
(ii)~We diagnose why our first version, based on posterior means, failed its pre-registered confirmation, and we fix it
with two standard tools. (iii)~We report three pre-registered rounds on up to 14 datasets, namely the failed test of the
first version, the confirmation of the corrected rule against \Cost, and a test of two choices that it inherits from \Cost.
(iv)~We give an implementation with the decision in C++, which makes the same decisions as the reference, and with an
automatic early end of the move phase. We also run the rule unchanged inside GPU e-means and o-means.

\paragraph{Revival as a Decision.} A move frees center $j$ and splits cluster $s$ (\cref{fig:teaser}). Its \emph{net gain} is the split gain
of $s$ minus the removal cost of $j$, and we make a move only if, given the evidence, the gain is more likely than not to
exceed the cost. Each center's estimates (of the rate at which its cluster receives points, of its removal cost per
point and of its split gain) are shrunk toward the population of centers by empirical
Bayes~\citep{morris1983parametric}, each with its own model (\cref{sec:estimation}). Thus, a center with little evidence is judged mostly by what is typical, and a center with much
evidence by its own data. As in \Cost, we choose moves by a greedy matching of the cheapest removals with the largest
gains. A newly split center waits until the weight of its inherited statistics has halved before it can be moved
again.

\paragraph{What Went Wrong First.} Our first version used posterior means and a moment estimate of the prior, and it
failed its pre-registered confirmation (\cref{sec:exp-round1}). Both choices let the centers with the least evidence
decide (\cref{sec:why}). First, the posterior mean credits a center without evidence with the prior mean of the gain,
which is large when the gains are spread out, so with five epochs empty centers became the best donors. Second, a moment
estimate of the prior's spread is dominated by the low-evidence centers, and on one dataset it shrank every gain to the
same value. Two standard tools fix both failures, namely the \emph{posterior median}, which gives no credit for ignorance, and the precision-weighted \emph{Paule--Mandel}
estimate of the prior~\citep{paule1982consensus}, in which centers without evidence barely count.

\paragraph{Evidence.} In a second pre-registered round on 14 datasets, 34 dataset--size combinations, and 20 seeds that
no earlier run had used, \Ours{} lowers the test and training errors of \Cost{} by 2.9\% and 8.9\%, and with five epochs
by 5.0\% and 7.1\% (\cref{sec:exp-round2}). It also has losses. With k-means++ starts, its test error is 2.4\% higher
than that of \Cost, all of it on one dataset whose test error is dominated by a single test point far from all training
data. On test error, it is not better than \Cost{} combined with our earlier distance test. A third round (\cref{sec:exp-round3}) shows that checking every five batches,
as \Cost{} does, gives lower errors than checking less often and errors within about 1\% of checking more often, which
costs more time. On the current revision of the pull request, dropping its \emph{freeze} (i.e., no moves in the last
pass) lowers both errors at two epochs and leaves them within 1\% at five. Our released implementation, which drops the
freeze by default, reproduces these results (\cref{sec:exp-final}). We report all three rounds in full.

\paragraph{Cost and Portability.} To \Cost, the rule adds a few per-point sums and, at every revival check, a search for
each center's nearest other center. With the decision in C++, which makes the same decisions as the reference
implementation in every check we compared, the rule takes a median 7\% longer than \Cost{} at two epochs, both on the
revision of our first two rounds and, without the freeze, on the current one. Because the few moves made late in
training are rarely worth their cost, we also give an automatic early end of the move phase, the \emph{early stop},
which saves about 5\% of the running time with five and ten epochs at a cost of about 0.1\% in test error and 0.3\% in
training error (\cref{sec:early-stop}). The decision is a function of per-center statistics only. It runs unchanged
inside the GPU implementation of e-means and inside o-means~\citep{bajaj2026omeans}, a variant of e-means with a
two-level assignment, where it lowers the test error of o-means' own count rule by a third (\cref{sec:exp-other}).
Optimized GPU versions take 30 to 70\% longer than the GPU rules they replace on runs of under half a second, and in
e-means this overhead falls as the data grow.

We release the code, the protocols, and the records of every run.\footnote{\raggedright\url{https://github.com/eyal-weiss/emeans-net-gain-revival}, archived on Zenodo
(\href{https://doi.org/10.5281/zenodo.24389631}{doi:10.5281/zenodo.24389631}).} A companion
technical report describes every alternative we tried and why it was dropped~\citep{weiss2026report}.

\Cref{sec:background} reviews e-means and its revival rules. \Cref{sec:method} defines the net gain of a move,
\cref{sec:estimation} shows how the rule estimates it, and \cref{sec:implementation} describes its implementation and
the early stop. \Cref{sec:experiments} reports the three rounds and the exploratory runs, \cref{sec:related} reviews
related work, \cref{sec:limitations} discusses the limitations, and \cref{sec:conclusion} concludes.

\section{Background: e-means and Its Revival Rules}\label{sec:background}

This section fixes the notation, describes the e-means update and its revival step, and reviews the four rules with
which we compare \Ours.

\paragraph{Notation.} The training points are $x_1,\dots,x_n\in\R^d$, the centers are $c_1,\dots,c_k$, and
$\nearest(x)=\argmin_j\lVert x-c_j\rVert^2$. The cluster of center $j$ is the set of points nearest to it. For a point
$x$ of cluster $j$, we write $d_1(x)=\lVert x-c_j\rVert^2$ and $d_2(x)=\min_{l\ne j}\lVert x-c_l\rVert^2$, and we call
$m(x)=d_2(x)-d_1(x)\ge 0$ its \emph{margin}, i.e., the increase in its squared error if $c_j$ disappeared. We write
$\delta_j=\min_{l\ne j}\lVert c_j-c_l\rVert^2$ for the squared distance from $c_j$ to its nearest other center $n(j)$.
The test and training errors are the mean squared distances from the test and training points to their nearest center.

\paragraph{The e-means Update.} e-means~\citep{bessa2026emeans} makes $T$ passes (epochs) over the data in batches of $B$
points. For each center, it keeps an EMA of the number of points that the center receives per batch, $N_j$, and an EMA
of their sum. After batch $t$, both are multiplied by $1-\alpha_t$, the batch's contribution is added with weight
$\alpha_t$, and the center is their ratio. We call $\alpha_t$ the step size and $N_j$ the \emph{usage} of center $j$.
The batch size $B$ and the initial step size $\alpha_0$ follow scaling laws in $n$, $k$ and $T$, and $\alpha_t$
decreases along a cosine schedule to a tenth of $\alpha_0$. After the last pass, each center is replaced by the mean of
the points it received in that pass.

\paragraph{Revival.} Every five batches, e-means runs a \emph{revival check}. A check frees some centers and chooses a
\emph{donor} cluster $s$ for each of them. The freed center is moved next to $c_s$, and the two become the two halves of
$s$. We call such a pair a \emph{move} $(j,s)$. The rules differ in which centers they free, which donors they choose,
and where they place the two halves.

\paragraph{\Lib{} (FAISS 1.16).} A center is freed when $N_j<0.1\,\overline N$, where $\overline N$ is the average usage.
Donors are drawn with probability proportional to usage, and the two halves are placed at $c_s$ plus and minus a small
random perturbation.

\paragraph{\Dist{}~\citep{weiss2026small}.} As \Lib, but a candidate is freed only if $\delta_j$ is at most four times
the average squared error per point. Thus, a small cluster far from every other center is kept, because its points
would be far from their next center.

\paragraph{\Cost{} (FAISS Pull Request \#5915).} While moves are allowed, every point's two nearest centers are found,
and each center keeps an EMA $R_j$ of the sum of the margins of its points per batch. This is an estimate of its removal
cost per batch in the sense of Fritzke's utility~\citep{fritzke1997lbgu}. Each center also keeps an EMA $U_s\in\R^d$ of
$\sum p(x)\,(x-c_s)$ over the points of cluster $s$, where $p(x)=\langle x-c_s,u_s\rangle$ is the projection on the
cluster's split axis $u_s=U_s/\lVert U_s\rVert$. This is Oja's rule for the cluster's principal
direction~\citep{oja1982simplified}, and $\lVert U_s\rVert$ estimates the cluster's scatter along it per batch. If the
projections of the points of $s$ on that axis were Gaussian with standard deviation $\sigma_s$, splitting $c_s$ into
$c_s\pm\sigma_s\sqrt{2/\pi}\,u_s$ would lower the squared error of its points by $G_s=\tfrac2\pi\lVert U_s\rVert$ per
batch, i.e., by the \emph{split gain}. A check frees the centers with $R_j<0.3\,\mathrm{median}(R)$, from the cheapest, and
pairs them with the donors of largest $G_s$. A freed center's nearest other center may not be freed at the same check (a
\emph{pair guard}). No move is made once less than one full pass of training remains, so that every final center has a
full pass of points (the \emph{freeze}). We call the part of training in which moves are allowed the \emph{move phase}. The pull
request also contains the training loop into which the rules plug, which we call the \emph{driver}. Our first two rounds
use the pull request at commit 98deaca. Its current revision, 79440bc, drops the freeze. Moves then continue to the end,
and a closing step reassigns the points of the last pass that a moved center held (\cref{sec:exp-head}). The third
round and our released implementation use this revision. The pull request was merged into FAISS on 22 November 2026,
after our runs; the commits after 79440bc change the rule only in how the GPU driver orders donors with NaN statistics.

\paragraph{\Hyb.} This rule is \Cost{} with the distance test of \Dist{} added to its removal decision. We built it while
comparing the two rules and include it as a strong baseline.

\section{The Net Gain of a Move}\label{sec:method}

In this section, we define what a move is worth, and in \cref{sec:estimation} we explain how the rule estimates it.
Throughout, costs and gains are changes in the total squared error per batch of $B$ points, and $\lambda_j$ is the
expected number of points per batch that fall in cluster $j$, called its \emph{rate}.

\paragraph{Removal Cost.} Freeing center $j$ while the other centers stay in place sends each point of its cluster to its
second-nearest center, which raises the squared error by the point's margin. Per batch, the cost is
\begin{equation}\label{eq:cost}
  C_j=\lambda_j\,\mu_j,\qquad \mu_j=\E[m(x)\mid x\text{ in cluster }j],
\end{equation}
which is Fritzke's utility of the center~\citep{fritzke1997lbgu}. When $c_j$ is the mean of its points, sending all of
them to the single center $c_{n(j)}$ would cost exactly $\delta_j$ per point (the parallel-axis identity), and letting
each point choose its own second-nearest center can only cost less. Hence $\mu_j\le\delta_j$, and we write
\begin{equation}\label{eq:ratio}
  \mu_j=r_j\,\delta_j,\qquad 0\le r_j\le 1\ \text{ when $c_j$ is the mean of its points}.
\end{equation}
The \emph{margin ratio} $r_j$ separates what the geometry of the centers tells us exactly ($\delta_j$, which carries
the isolation of $c_j$) from what must be estimated from points ($r_j$, i.e., the share of that distance that the points
incur). \Dist{} uses $\delta_j$ alone, whereas \Cost{} estimates $\lambda_j\mu_j$ directly.

\paragraph{Split Gain.} We take the split gain of \Cost, which splits $c_s$ into $c_s\pm a_su_s$, with the axis
$u_s=U_s/\lVert U_s\rVert$ and the offset $a_s=\sqrt{2\lVert U_s\rVert/(\pi N_s)}$. For a fixed offset $a$, sending each
point of $s$ to the nearer half lowers its squared error by $2a\lvert p(x)\rvert-a^2$. Thus, the cluster's squared error
per batch falls by $\lambda_s(2a\,\E\lvert p\rvert-a^2)$. If the projections $p$ are centered Gaussian with variance
$\sigma_s^2$, this decrease is largest at $a=\sigma_s\sqrt{2/\pi}$, where it equals $G_s=\tfrac2\pi\lambda_s\sigma_s^2$.
If $\lVert U_s\rVert$ and $N_s$ estimate the scatter along the axis per batch, $\lambda_s\sigma_s^2$, and the rate
$\lambda_s$, then $a_s$ is that offset and $\tfrac2\pi\lVert U_s\rVert$ estimates $G_s$. Under these idealized
conditions, $G_s$ is the gain of the split as placed, with each point choosing between the two halves. Points that may
also move to other centers can only gain more. We also tried measuring the gain directly from the points, which made the
rule worse (the technical report~\citep{weiss2026report} explains why).

\paragraph{Net Gain.} The rule values a move $(j,s)$ by
\begin{equation}\label{eq:net}
  \Delta_{js}=G_s-C_j .
\end{equation}
This is a \emph{separable surrogate}, i.e., it ignores that the two changes interact when points of $j$ would have gone to
$s$, or points of $s$ to $j$, and that moves of the same check interact. It also ignores the later adaptation of all
centers, which usually makes a split pay more than its immediate effect. Separability lets the rule rank removals and
splits separately, which is what makes the matching below cheap.

\paragraph{Matching.} At a check, the rule sorts the eligible centers by estimated cost $\tilde C_j$, from the cheapest,
and the eligible donors by estimated gain $\tilde G_s$, from the largest (\cref{alg:check},
Lines~\ref{line:donors}--\ref{line:cands}). It visits the candidates in order and pairs each with the best donor that is
still available and other than itself (Line~\ref{line:pick}). It accepts the pair if $\tilde C_j<\tilde G_s$. It stops
at the first failed pair, unless the candidate failed only because the better donor was the candidate itself
(Line~\ref{line:test}). \Cref{app:spec} gives the exact procedure. The rule keeps the pair guard of \Cost{}
(Line~\ref{line:guard}). That is, a freed center's nearest other center is protected and may not be freed at the same
check, and a candidate whose own nearest other center has already been freed is skipped unless the two coincide exactly.
The next result shows that, when no candidate is blocked, the matching is optimal for the surrogate.

\begin{proposition}[Optimal Greedy Matching]\label{prop:greedy}
Sort costs increasingly, $\tilde C_{(1)}\le\tilde C_{(2)}\le\dots$, and gains decreasingly,
$\tilde G_{(1)}\ge\tilde G_{(2)}\ge\dots$, and let $m^*$ be the number of $i$ with $\tilde G_{(i)}>\tilde C_{(i)}$. If
the $m^*$ cheapest candidates and the $m^*$ largest-gain donors are disjoint, the pair guard allows the $m^*$ cheapest
candidates together, and every pairing between the two sets is allowed, then the greedy matching maximizes the total
surrogate value $\sum(\tilde G_s-\tilde C_j)$ over all sets of disjoint allowed moves.
\end{proposition}
In other words, when the guard does not interfere, no set of disjoint allowed moves has a larger total predicted net
gain than the set that the greedy matching makes.
\begin{proof}
For a set of $m$ moves, the total value is the sum of the gains of its $m$ donors minus the sum of the costs of its $m$
freed centers, whatever the pairing. It is therefore at most the sum of the $m$ largest gains minus the sum of the $m$
smallest costs, $V(m)=\sum_{i\le m}(\tilde G_{(i)}-\tilde C_{(i)})$, whether or not those sets are feasible. The terms
$\tilde G_{(i)}-\tilde C_{(i)}$ are non-increasing in $i$, so $V(m)$ increases while the terms are positive and
decreases afterwards. Hence its maximum over $m$ is $V(m^*)$. Under the assumptions, the greedy matching pairs exactly
the $m^*$ cheapest candidates with the $m^*$ largest gains and stops, since the next term is not positive. Thus, it
attains $V(m^*)$.
\end{proof}
When the guard blocks a candidate, the greedy matching can be suboptimal (\cref{app:spec} gives an example). We keep the
guard because it prevents freeing both of two nearby centers, which the surrogate cannot see.

\paragraph{Cool-Down.} After a split, the donor and the freed center start from the donor's running statistics (its
usage and split statistics), halved, and their statistics since the split start from zero. While the inherited part
dominates, the estimates describe the old cluster, and acting on them moves centers back and forth. The rule therefore
lets neither of the two be freed or used as a donor while the coefficient of the inherited contribution,
$h=\prod(1-\alpha_t)$ over the batches since the split, exceeds one half, i.e., for one half-life of the EMA. This halves
the inherited coefficient. Note that it does not guarantee that half of a center's statistics are its own (a center that receives
no points keeps only inherited statistics), but it lets such a center become eligible again. The half-life is long late
in training, when $\alpha_t$ is small, and short early on. In our development runs, the cool-down removed every move made
on mostly inherited statistics, and it cut the number of moves by a third to three quarters on the datasets where moves
were frequent.

\paragraph{Everything Else Is \Cost.} The rule changes only which moves are made. The placement of the two halves, the
check every five batches (a choice that the rule does not derive; \cref{sec:exp-round3}) and the timing of moves are
those of \Cost{} and its driver. On the current revision of the driver, our implementation drops the freeze by default,
which lowers the errors at two epochs (\cref{sec:exp-head}), and keeps it as an option.

\begin{algorithm}[t]
\caption{One revival check of \Ours. Color marks what differs from \Cost, and a bar in the margin marks the lines that contain it.}\label{alg:check}
\begin{algorithmic}[1]
\Require \cbstart per center: usage $N_j$, \newpart{points $n_j$ and batches $a_j$ since its birth, margin sums $M_j,Q_j$,
  projection sums $P^{(2)}_j,P^{(4)}_j$, inherited coefficient $h_j$}; split statistics $U_j$\cbend
\Ensure the moves $(j,s)$ made at this check
\State \cbstart\newpart{$\delta_j,n(j)\gets$ squared distance to, and index of, the nearest other center, for every $j$}
\State \newpart{$\tilde\lambda_j\gets$ posterior mean rate (Gamma--Poisson, \cref{eq:rate})}
\State \newpart{$\tilde r_j\gets$ shrunk margin ratio (\cref{eq:rhat}); \quad $\tilde C_j\gets\tilde\lambda_j\,\tilde r_j\,\delta_j$}
\State \newpart{$\tilde G_s\gets$ posterior median of the split gain (\cref{eq:gain-post})}
\State donors $D$: centers with $N_s>0$ \newpart{and $h_s\le\tfrac12$}, sorted by \newpart{$\tilde G_s$} decreasing\label{line:donors}
\State candidates: centers with \newpart{$h_j\le\tfrac12$ and $\tilde C_j<\tilde G$ of the first donor}, sorted by
  \newpart{$\tilde C_j$} increasing\label{line:cands}\cbend
\For{each candidate $j$ in order}
  \State $s\gets$ the first donor in $D$ that is not freed, not used, and $\ne j$\label{line:pick}
  \If{there is none} \textbf{stop} if every donor is freed or used, else go to the next candidate \EndIf
  \If{\cbstart\newpart{$\tilde C_j\ge\tilde G_s$}}\label{line:test} \textbf{stop} if $s$ is the first donor not freed or used, else go to the
  next candidate\cbend \EndIf
  \If{$j$ is protected, or was used as a donor, or $n(j)$ is freed and $c_j\ne c_{n(j)}$}\label{line:guard} go to the next candidate \EndIf
  \State \cbstart move $(j,s)$: free $j$, protect $n(j)$, use $s$; split $c_s$ along $u_s$; \newpart{restart the statistics since
  birth of $j$ and $s$; $h_j,h_s\gets1$}\cbend
\EndFor
\end{algorithmic}
\end{algorithm}

\section{Estimation Under Unequal Evidence}\label{sec:estimation}

In this section, we describe how the rule estimates $C_j=\lambda_jr_j\delta_j$ and $G_s$ for every center at every
check. Only $\delta_j$ is known exactly. The rest must be estimated from the points each center has received, and the
amount of evidence varies by orders of magnitude between centers. We call the start of training, or the split that
created or last split a center, its \emph{birth}. Since its birth, the rule keeps for each center the number of batches
$a_j$ and the number of points $n_j$. Over those points, it keeps the sums of the margins $M_j=\sum m(x)$ and of their squares
$Q_j$, and the sums $P^{(2)}_j$ and $P^{(4)}_j$ of the second and fourth powers of the projections on its split axis.
While moves are allowed, the two-nearest search of \Cost{} already runs on every point, so these sums cost a few
operations per point.

Every estimate follows the same pattern. It combines a raw estimate from the center's own points, an estimate of its
noise, and a \emph{prior} fitted to the population of all centers. The estimate used is a compromise between the center's raw
estimate and the population, weighted by their precisions~\citep{morris1983parametric}. Thus, a center with much
evidence keeps its own estimate, and a center with little evidence gets what is typical.

\paragraph{Rates (Gamma--Poisson).} Suppose that the points a center receives per batch are Poisson with rate
$\lambda_j$, and that the rates of the centers follow a Gamma distribution with shape $a$ and rate $b$. Then, given $n_j$
points in $a_j$ batches, the posterior mean is
\begin{equation}\label{eq:rate}
  \tilde\lambda_j=\frac{a+n_j}{b+a_j}.
\end{equation}
The prior is fitted by moments over the centers with $a_j\ge1$. Let $\bar\lambda$ be the mean of $n_j/a_j$, and let $v$
be the variance of $n_j/a_j$ minus its Poisson part, i.e., the mean of $n_j/a_j^2$. Then $a=\bar\lambda^2/v$ and
$b=\bar\lambda/v$. A center born at this check gets the prior mean $a/b$.

\paragraph{Margin Ratios (Normal--Normal).} The raw ratio is $\hat r_j=M_j/(n_j\delta_j)$. Its noise variance is
$\sigma^2/n_j$, where $\sigma^2$ is the variance of $m(x)/\delta_j$ within a cluster, pooled over the clusters with at
least two points. The ratios are shrunk toward their population mean $\bar r$ (weighted by points),
\begin{equation}\label{eq:rhat}
  \tilde r_j=\frac{\tau_r^2\,\hat r_j+(\sigma^2/n_j)\,\bar r}{\tau_r^2+\sigma^2/n_j},
\end{equation}
where $\tau_r^2$, the variance of the true ratios between clusters, is the variance of the $\hat r_j$ minus the average
noise variance. The estimated cost is $\tilde C_j=\tilde\lambda_j\tilde r_j\delta_j$.

\paragraph{Split Gains (Log Scale).} The raw gain $\hat G_s=\tfrac2\pi\lVert U_s\rVert/(1-\prod_t(1-\alpha_t))$ removes
the start-up bias of an EMA started at zero. Gains vary over orders of magnitude between clusters, and their noise is
multiplicative. The rule therefore works with $y_s=\log\hat G_s$. As a working approximation, we treat $\lVert
U_s\rVert$ like a sum of squared projections over a fixed number $n^{\mathrm{eff}}_s=\tilde\lambda_sW_t$ of points. Here
$W_t=\min\{(2-\alpha_t)/\alpha_t,\,t+1\}$ approximates the memory of the EMA in batches, and $t$ is the zero-based index
of the batch. The relative variance of such a sum is $(\kappa-1)/n^{\mathrm{eff}}_s$, where $\kappa$ is the kurtosis of
the projections ($\kappa=3$ for Gaussian data). To first order (the delta method), this is the variance of its
logarithm. The rule thus takes $v_s=(\bar\kappa-1)/\max(n^{\mathrm{eff}}_s,10^{-3})$, where $\bar\kappa$ is the kurtosis
pooled over all clusters (weighted by points), because the kurtosis of one cluster is itself too noisy. This ignores the
variability of the number of points. A Poisson number of points would give $\bar\kappa/n^{\mathrm{eff}}_s$, and that
version did worse in our tests~\citep{weiss2026report}. With $\theta_s=\log G_s$ the true log gain, the working model is
$y_s\mid\theta_s\sim\mathcal N(\theta_s,v_s)$ and $\theta_s\sim\mathcal N(\mu,\tau^2)$. The posterior of $\theta_s$ is
then normal with mean
\begin{equation}\label{eq:gain-post}
  \tilde y_s=\frac{\tau^2y_s+v_s\mu}{\tau^2+v_s},\qquad \tilde G_s=\exp(\tilde y_s).
\end{equation}

\paragraph{The Prior (Paule--Mandel).} The mean and spread of the prior are fitted to the population by the
Paule--Mandel estimator~\citep{paule1982consensus}, the standard choice for this random-effects model in
meta-analysis~\citep{veroniki2016methods}. With weights $w_s=1/(\tau^2+v_s)$, $\mu$ is the weighted mean of the $y_s$,
and $\tau^2$ is the value at which the weighted spread $\sum_sw_s(y_s-\mu)^2$ equals its expected value $m-1$, where $m$
is the number of centers with a positive gain. If the spread is below $m-1$ already at $\tau^2=0$, then $\tau^2=0$. A
center with a large noise variance has a small weight and barely moves $\mu$ or $\tau^2$. \Cref{app:spec} gives the
numerical details of the fit.

\paragraph{The Decision Uses Medians.} Under the working model, with the fitted $\mu$ and $\tau^2$ treated as known,
$\tilde G_s=\exp(\tilde y_s)$ is the posterior \emph{median} of the gain, i.e., the gain exceeds it with posterior
probability one half. The cost is a posterior mean (of the rate, times the shrunk ratio). The decision $\tilde
C_j<\tilde G_s$ therefore means that \emph{the gain of the split is more likely than not to exceed the estimated cost of the
removal}. Note that this is not a statement about the joint posterior of gain and cost. We also tried replacing the posterior mean
of the cost by its median, and in exploratory runs the training error rose by up to 7\%~\citep{weiss2026report}.

\subsection{Why the Posterior Mean and a Moment Prior Fail}\label{sec:why}

Our first rule, \Mean, used the posterior mean of the gain, $\exp(\tilde y_s+\tilde v_s/2)$ with
$\tilde v_s=\tau^2v_s/(\tau^2+v_s)$, and the moment estimate $\tau^2=\mathrm{Var}(y)-\overline{v}$, with $\mu$ the plain
mean of the $y_s$. Both choices let the centers with the least evidence decide, in two different ways.

\paragraph{The Mean Rewards Ignorance.} Consider the idealized limit of no evidence, $v_s\to\infty$ for a center with a
positive raw gain, with $\mu$ and $\tau^2$ fixed. Then $\tilde y_s\to\mu$ and $\tilde v_s\to\tau^2$, and the posterior
mean of its gain tends to $\exp(\mu+\tau^2/2)$, the prior mean of the gain, which is large when the gains of the
population are spread out. (The implementation floors $n^{\mathrm{eff}}$ at $10^{-3}$, so $v_s$ stays finite, and a
center with zero raw gain keeps gain zero.) With five epochs, the step size starts near one. The EMAs of a center that
stops receiving points thus decay within a few batches to raw gains of $10^{-7}$ to $10^{-11}$, which spread the log
gains of the population. The median $\tau^2$ over the checks of a pass rose from 32 in the first pass to between 98 and
182 in the later passes (\cref{tab:runaway}). A center without points was then credited with a gain of $10^{13}$ to
$10^{18}$ and chosen as the best donor, and splitting it created two more empty centers. In passes 3 and~4, 81--84\% of
all moves split a donor that had received no point in the last five batches, and the runs ended with 537 to 761 of
their 1{,}024 centers empty. The median gives such a center $\exp(\mu)$, a typical gain, and the runaway disappears.

\paragraph{A Moment Prior Is Decided by the Noisiest Centers.} The moment estimate of $\tau^2$ subtracts the
\emph{average} noise variance, which is dominated by the centers with the least evidence. On KDD Cup~99 with $k=256$,
the pooled $\bar\kappa-1$ was 55 to 266, so $v_s$ was enormous for low-evidence centers. The subtraction drove $\tau^2$
to its floor at four of the first six checks, and every gain was shrunk to the same value, a \emph{collapse} of the
prior. Hence the center that had absorbed a large unexplored region could not be recognized as the best donor, as its
raw gain of $1.85\cdot10^5$ became~$13.7$. That run ended with a test error almost ten times that of \Cost. The
Paule--Mandel estimate weights each center by its precision, so the noisy centers barely count, and the same run ends
level with \Cost.

\Cref{tab:ablation} separates the two fixes. The median alone stops the runaway but not the collapse, and the
Paule--Mandel prior alone removes most of the collapse but not the runaway. Together, they match or beat \Cost.

\begin{table}[t]
\centering
\caption{The runaway of \Mean{} with five epochs on KDD Cup 99, $k=1{,}024$ (development seeds 10--14, totals over
seeds). The columns give the moves per pass, the share of moves whose donor had received no point in the last five
batches, the median of the prior spread $\tau^2$ over the checks of the pass, and the medians of the donor's raw and
posterior-mean gains over the moves whose donor had received no point in the last five batches.}\label{tab:runaway}
\footnotesize
\begin{tabular}{rrrrrr}
\toprule
Pass & Moves & Donor without points & $\tau^2$ & Raw gain & Posterior-mean gain\\
\midrule
1 & 4{,}032 & 6\% & 32 & $1.8\cdot10^{-7}$ & $1.1\cdot10^{5}$\\
2 & 3{,}669 & 56\% & 98 & $5.1\cdot10^{-8}$ & $1.8\cdot10^{13}$\\
3 & 6{,}336 & 84\% & 182 & $3.4\cdot10^{-10}$ & $3.4\cdot10^{18}$\\
4 & 3{,}756 & 81\% & 123 & $5.9\cdot10^{-11}$ & $8.5\cdot10^{15}$\\
\midrule
\multicolumn{6}{l}{For comparison, two epochs, pass 1: 4{,}309 moves, 4\%, $\tau^2=2.4$}\\
\bottomrule
\end{tabular}
\end{table}

\begin{table}[t]
\centering
\caption{The two fixes separately, as the test error on KDD Cup 99 with development seeds, with \Cost{} for reference (a range
is over the seeds of the row). Mean or median: the posterior summary of the gain. Moments or P--M: the moment or the
Paule--Mandel estimate of the prior.}\label{tab:ablation}
\footnotesize
\setlength{\tabcolsep}{4pt}
\begin{tabular}{lccc>{\columncolor{nmtint!8}}cc}
\toprule
Case & \stackhead{Mean, moments}{(\Mean)} & Median, moments & Mean, P--M & \stackhead{Median, P--M}{(\Ours)} & \Cost\\
\midrule
$k=1024$, 5 epochs, seeds 10--14 & 1.09--18.7 & 0.49--0.51 & 5.33--25.1 & 0.49--0.51 & 0.49--0.53\\
$k=256$, 2 epochs, seed 13 & 17.3 & 17.7 & 2.72 & 1.79 & 1.80\\
\bottomrule
\end{tabular}
\end{table}

\section{Implementation and Cost}\label{sec:implementation}

In this section, we describe what the rule adds to the running time of \Cost, how we wrote its decision in C++, and the
early stop.

\paragraph{What the Rule Adds to \Cost.} \Cost{} already finds every point's two nearest centers while moves are
allowed, and it keeps the split statistics $U_s$. Per point, the rule adds the point's margin and its projection on its
cluster's split axis (a dot product), accumulated into five sums per center. Per check, it adds the search for every
center's nearest other center ($\delta_j$, $n(j)$) and the decision itself, which costs $O(k\log k)$. The
nearest-other-center search compares $k$ centers with $k$ centers every five batches, while the assignment compares $B$
points with $k$ centers every batch. The search thus adds about $k/(5B)$ of the assignment time during the move phase.
With the freeze at $T=2$, the move phase is the first of the two passes, so the share of the whole run is about
$k/(5TB)$. Without the freeze, the move phase lasts until the early stop (\cref{sec:early-stop}) or the end. Measured on
one machine, the search added 1.3\% on KDD Cup~99 with $k=4{,}096$, 4.7\% on DEEP~1M with $k=16{,}384$, and 10\% on
GeoNames with $k=4{,}096$, within a few tenths of a percent of the formula. On data of two or three dimensions, a k-d
tree would make it cheaper. In high dimensions, an approximate search with an inverted file (IVF) halved its cost at
$k=4{,}096$ without changing the results in our tests.

\paragraph{The Decision in C++.} We moved the whole decision from NumPy to C++, and it makes the same decisions as the
reference implementation in every check we compared (\cref{app:cpp}). The decision became three to six times faster,
and the median extra running time of the rule relative to \Cost{} fell from 10\% to about 7\% (from 31\% to 20\% on the
small Shuttle dataset, where the decision mattered most).

\subsection{An Automatic Early End of the Move Phase}\label{sec:early-stop}

Most moves happen early. In our logged runs on the low-dimensional datasets (household power, 3D Road, GeoNames,
cod-rna), 78--94\% of all moves were made in the first tenth of the move phase. Afterwards, the checks, the two-nearest
search and the extra per-point sums keep running and rarely change anything. On KDD Cup~99, Shuttle and
YearPredictionMSD, in contrast, moves continue until the freeze.

We therefore let the rule end the move phase by itself (the early stop). A check is \emph{quiet} if it accepts no move.
A clock sums $\log(1-\alpha_t)$ over the batches since the start of training and is reset to zero at every check that
is not quiet. The move phase ends at a quiet check once the clock has reached $\log\frac12$. At that point, one
half-life of the EMA has passed since the last check that was not quiet (or since the start), so the coefficient of
everything the EMA held then has halved, and still no move qualified. The moves accepted at the stopping check, if any,
are made. From then on, the rule adds nothing, i.e., no extra statistics and no checks. (On the earlier revision of the
driver, the algorithm then runs as after the freeze of \Cost, with a plain nearest-center search.) Note that the early
stop uses the same half-life as the cool-down and adds no parameter. A variant also counts a check as quiet when the predicted net
gain of its accepted moves, $\sum(\tilde G_s-\tilde C_j)$, is strictly below $10^{-4}$ times the squared error of the
last batch. A check with a few small moves can then be quiet, and the clock keeps running. This variant stops earlier
and saves more, at a larger cost in error (\cref{sec:exp-stop}).

\section{Experiments}\label{sec:experiments}

The analysis of \cref{sec:method,sec:estimation} does not tell whether the rule lowers the error in practice, at what
cost in running time, and whether it stays stable on longer schedules. We therefore tested it in three pre-registered
rounds, of \Mean{} (\cref{sec:exp-round1}), of \Ours{} (\cref{sec:exp-round2}), and of two choices that \Ours{} takes over
from \Cost{} (\cref{sec:exp-round3}). A final comparison of our released implementation tests no new hypothesis
(\cref{sec:exp-final}). Everything else after the second round (the early stop, GPU e-means and o-means) is
exploratory and labeled as such.

\subsection{Protocol}\label{sec:protocol}

\paragraph{Data.} We used fourteen datasets (\cref{tab:datasets}). Six are those of our earlier
study~\citep{weiss2026small}, namely KDD Cup 99~\citep{kddcup99}, the 1990 US Census~\citep{census1990},
YearPredictionMSD~\citep{bertin2011msd}, GeoNames~\citep{geonames}, household power~\citep{uci_power} and
Shuttle~\citep{uci_shuttle}. Eight were added between the rounds by a rule fixed before any run (public, dense, numeric,
at least 60{,}000 distinct rows, not used before). They are MNIST~\citep{lecun1998mnist},
Fashion-MNIST~\citep{xiao2017fashion}, Corel~\citep{uci_corel}, 3D Road~\citep{kaul2013road},
cod-rna~\citep{uzilov2006codrna}, IJCNN1~\citep{prokhorov2001ijcnn}, SensIT~\citep{duarte2004sensit} and a subsample of
HIGGS~\citep{baldi2014higgs}. We removed duplicates, held out 10{,}000 rows as the test set, and standardized the
features with training statistics (images were divided by 255). With two or three values of $k$ per dataset, there are
34 dataset--size combinations, which we call \emph{cells}.

\paragraph{Runs.} Every run trained e-means with the batch size and step schedule of the library, starting from $k$
distinct random training rows, for $T=2$ epochs unless stated otherwise. The rules differ only in revival. All rules ran
on the driver of the pull request at commit 98deaca, so every difference comes from the revival decision. Runs that are
compared share their data, their start and their machine.

\paragraph{Hardware.} The CPU runs used two shared servers. One has an AMD Ryzen Threadripper 3990X processor (64
cores, without AVX-512), 188\,GiB of memory and Ubuntu~24.04, and the other a Threadripper 7980X (64 cores, with
AVX-512), 251\,GiB of memory and Ubuntu~20.04. Each run used four threads, and each server ran up to eight runs at a
time. In rounds 2 and 3 and in the final comparison, the first server ran KDD Cup~99, GeoNames, household power and
Shuttle, and the second the other ten datasets. Both used Python~3.12, NumPy~2.5 with OpenBLAS~0.3.34, and FAISS~1.16.
The GPU runs used one NVIDIA RTX 3090 Ti (24\,GB) in the first server.

\paragraph{Statistics.} The unit of inference is the seed. For two rules and an endpoint (test or training error), we
computed, for every seed, the mean over the cells of the log ratio of the two errors. We report the geometric-mean ratio
over seeds minus one, the Hodges--Lehmann pseudomedian, and a two-sided Wilcoxon signed-rank test over the
seeds~\citep{wilcoxon1945}. Families of tests are corrected by Holm's method~\citep{holm1979}. A result counts only if
the estimate and the pseudomedian agree in sign. We report both test and training error, and in the second round both
were primary. Indeed, on Shuttle and KDD Cup~99, the test error is dominated by a few test points far from all training
data. On Shuttle, one test point accounts for between about two thirds and nine tenths of a run's test error. The test
error there thus measures how often a center happens to lie near those points rather than how well the data are
summarized.

\paragraph{Implementation and Gates.} We ran the version with the per-point sums and the matching in C++. Its
results were identical to those of the pure Python reference in both rounds (\cref{app:protocol}).

\begin{table}[t]
\centering
\caption{Datasets and, for the primary part of round 2 (random starts, 2 epochs, 20 seeds), the change in test error (upper line) and training
error (lower line) relative to \Cost{} in percent, as the geometric mean over the dataset's cells and seeds. Negative is better. The
columns $n$, $d$ and $k$ give the size and dimension of each dataset and the numbers of centers used.}
\label{tab:datasets}
\footnotesize
\setlength{\tabcolsep}{3pt}
\begin{tabular}{@{}lrrlrrrr>{\columncolor{nmtint!8}}rr@{}}
\toprule
Dataset & $n$ & $d$ & $k$ & \Lib & \Dist & \stackhead{\textsc{cost+}}{\textsc{distance}} & \Mean & \Ours & \stackhead{\textsc{net-median}}{\textsc{-fh}}\\
\midrule
KDD Cup 99 & 1.05M & 122 & 256--4096 & +1393 & +516 & +2.2 & +14.3 & +1.6 & +2.7\\
 &  &  &  & +1280 & +382 & $-$53.1 & $-$33.9 & $-$39.0 & $-$46.3\\[1.5pt]
Census 1990 & 1.98M & 68 & 256--4096 & +5.1 & +5.0 & +0.0 & $-$0.8 & $-$0.8 & $-$0.8\\
 &  &  &  & +4.8 & +4.7 & +0.0 & $-$0.8 & $-$0.8 & $-$0.7\\[1.5pt]
YearPredictionMSD & 495k & 90 & 256--4096 & +0.7 & +0.8 & +1.3 & +0.6 & +0.5 & +0.5\\
 &  &  &  & +2.8 & +1.4 & +0.6 & +0.1 & +0.1 & +0.1\\[1.5pt]
GeoNames & 1.00M & 2 & 256--4096 & +94.7 & +62.6 & +2.7 & $-$1.3 & $-$1.3 & $-$1.2\\
 &  &  &  & +101 & +68.1 & +1.5 & $-$2.1 & $-$2.0 & $-$2.0\\[1.5pt]
household power & 1.89M & 7 & 256, 1024 & +37.8 & +34.2 & $-$0.4 & $-$2.7 & $-$2.6 & $-$2.6\\
 &  &  &  & +37.8 & +34.1 & +0.2 & $-$2.5 & $-$2.5 & $-$2.5\\[1.5pt]
Shuttle & 38k & 9 & 256, 1024 & +519 & +157 & $-$37.1 & $-$40.8 & $-$37.2 & $-$38.6\\
 &  &  &  & +1614 & +362 & $-$42.5 & $-$52.1 & $-$52.7 & $-$51.8\\[1.5pt]
MNIST & 50k & 784 & 256, 1024 & +0.4 & +0.4 & $-$0.0 & +0.1 & +0.1 & +0.1\\
 &  &  &  & +0.6 & +0.6 & +0.0 & +0.0 & +0.0 & +0.1\\[1.5pt]
Fashion-MNIST & 50k & 784 & 256, 1024 & +0.6 & +0.6 & +0.0 & +0.2 & +0.2 & +0.2\\
 &  &  &  & +0.9 & +0.9 & +0.0 & +0.1 & +0.1 & +0.1\\[1.5pt]
Corel & 48k & 16 & 256, 1024 & +12.8 & +7.9 & $-$0.2 & $-$0.4 & $-$0.3 & $-$0.6\\
 &  &  &  & +13.0 & +9.3 & $-$0.1 & $-$1.1 & $-$1.1 & $-$1.0\\[1.5pt]
3D Road & 415k & 3 & 256--4096 & +14.9 & +14.1 & +0.5 & $-$0.2 & $-$0.2 & $-$0.2\\
 &  &  &  & +16.0 & +15.0 & +0.4 & $-$0.3 & $-$0.2 & $-$0.1\\[1.5pt]
cod-rna & 323k & 8 & 256--4096 & +3.7 & +3.3 & +0.0 & $-$0.3 & $-$0.3 & $-$0.2\\
 &  &  &  & +4.0 & +3.6 & +0.0 & $-$0.3 & $-$0.3 & $-$0.3\\[1.5pt]
IJCNN1 & 122k & 22 & 256, 1024 & +7.9 & +6.9 & +0.2 & $-$0.6 & $-$0.5 & $-$0.5\\
 &  &  &  & +8.9 & +8.0 & +0.1 & $-$0.7 & $-$0.7 & $-$0.7\\[1.5pt]
SensIT & 79k & 100 & 256, 1024 & +1.0 & +1.3 & +0.6 & +0.5 & +0.5 & +0.5\\
 &  &  &  & +3.2 & +2.7 & +0.1 & $-$0.1 & $-$0.0 & +0.1\\[1.5pt]
HIGGS (subsample) & 78k & 28 & 256, 1024 & +0.2 & +0.1 & +0.1 & +0.1 & +0.1 & $-$0.0\\
 &  &  &  & +0.5 & +0.4 & +0.0 & +0.0 & +0.0 & +0.0\\
\bottomrule
\end{tabular}

\end{table}

\begin{figure}[t]
\centering
\includegraphics{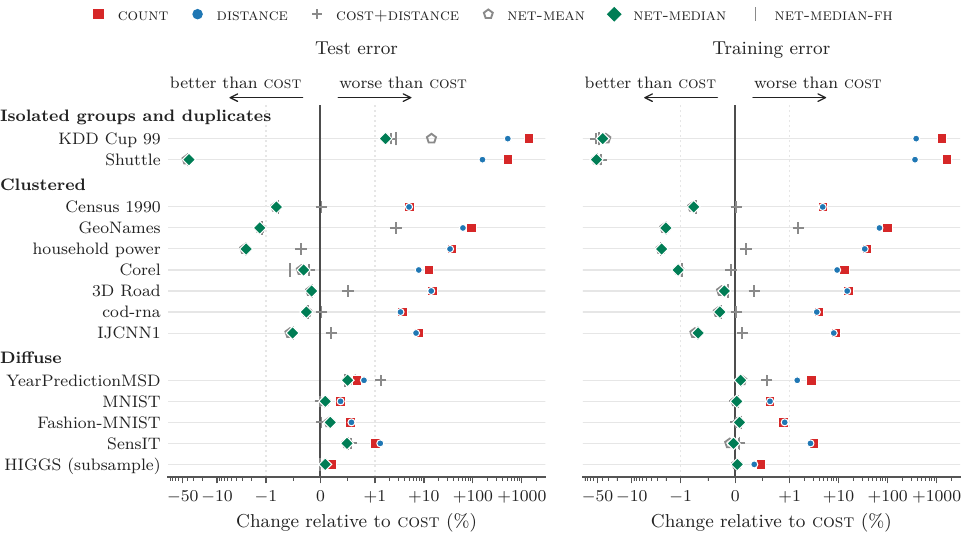}
\caption{Round 2, primary part (random starts, 2 epochs, 20 seeds): change in test error (left) and training error
(right) of each rule relative to \Cost{} per dataset, as the geometric mean over the dataset's cells and seeds (the
values of \cref{tab:datasets}). Negative is better; the vertical line at zero is \Cost. The scale is linear between
$-1\%$ and $+1\%$ (dotted lines) and logarithmic beyond. Rows are grouped as in \cref{sec:exp-round2}. \Ours{} (green
diamonds) is drawn above the other rules, and \Dist{} above \Lib.}\label{fig:round2}
\end{figure}

\subsection{Round 1: Posterior Means Are Not Enough}\label{sec:exp-round1}

The first round tested \Mean{} against \Cost{} on the 16 cells of the six original datasets, with 20 seeds that no
development run had used. Its primary hypothesis, a lower test error with random starts and two epochs, was not
confirmed (the estimate was $-2.2\%$ with $p=0.26$). It was better on four datasets (on Shuttle only with $k=256$) and worse
on two, most of all on KDD Cup~99 (\cref{app:details}). With k-means++ starts, its test error was 11\% higher than that of
\Cost. With five epochs, it broke down on KDD Cup~99 with $k=1{,}024$ in all five seeds tested, with 537 to 761 empty
centers at the end. Its training error, an exploratory endpoint in that round, was 19.6\% lower than that of \Cost{} (20
of 20 seeds).

\Cref{sec:why} traces both failures to the low-evidence centers. Before the second round, we also changed the evaluation
in three ways. Training error became a co-primary endpoint. Eight datasets were added, so that two unusual datasets
would not decide the outcome. MNIST and Fashion-MNIST were scaled by 255 rather than per pixel, since per-pixel scaling
had made a few test images artificial outliers. The protocol of the second round records all these changes.

\subsection{Round 2: Primary and Secondary Results}\label{sec:exp-round2}

The second round tested \Ours, with \OursFH{} as an alternative, on all 34 cells with seeds 30--49, which no run of the
project had used. The alternative differs only in the prior of the gain, whose mean depends on the center's rate and on
its distance to its nearest center (a Fay--Herriot model~\citep{fay1979estimates}, see \cref{app:spec}). Thus, a sparse
but distant region is not judged against the bulk of the data.

\begin{table}[t]
\centering
\caption{Round 2: pre-registered tests. Estimate: geometric-mean ratio minus one over seeds; pseudomedian in parentheses;
$p$ Holm-adjusted within each family (primary: 2 tests; secondary: 10); seeds: seeds on which the first rule had the
lower error. P: random starts, 2 epochs, 20 seeds; S5: 5 epochs, 10 seeds; SK: k-means++ starts, 10 seeds.}
\label{tab:round2}
\footnotesize
\begin{tabular}{llllrl}
\toprule
Part & Comparison & Endpoint & Estimate & $p$ & Seeds\\
\midrule
\multicolumn{6}{l}{\emph{Primary}}\\
P & \Ours{} vs \Cost & test & $-2.9\%$ ($-3.1\%$) & $6\cdot10^{-5}$ & 18/20\\
P & \Ours{} vs \Cost & training & $-8.9\%$ ($-8.7\%$) & $4\cdot10^{-6}$ & 20/20\\
\addlinespace[3pt]
\multicolumn{6}{l}{\emph{Secondary}}\\
P & \OursFH{} vs \Cost & test & $-2.9\%$ ($-3.4\%$) & 0.001 & 17/20\\
P & \OursFH{} vs \Cost & training & $-9.8\%$ ($-9.5\%$) & $2\cdot10^{-5}$ & 20/20\\
P & \Ours{} vs \Mean & test & $-0.7\%$ ($-0.5\%$) & 0.10 & 13/20\\
P & \Ours{} vs \Lib & test & $-38.5\%$ ($-38.5\%$) & $2\cdot10^{-5}$ & 20/20\\
P & \Ours{} vs \Dist & test & $-28.5\%$ ($-28.4\%$) & $2\cdot10^{-5}$ & 20/20\\
P & \Ours{} vs \Hyb & test & $-0.8\%$ ($-0.8\%$) & 0.10 & 13/20\\
\addlinespace[2pt]
S5 & \Ours{} vs \Cost & test & $-5.0\%$ ($-4.9\%$) & 0.012 & 10/10\\
S5 & \Ours{} vs \Cost & training & $-7.1\%$ ($-7.2\%$) & 0.012 & 10/10\\
\addlinespace[2pt]
SK & \Ours{} vs \Cost & test & $+2.4\%$ ($+2.1\%$) & 0.012 & 1/10\\
SK & \Ours{} vs \Cost & training & $-17.8\%$ ($-17.8\%$) & 0.012 & 10/10\\
\bottomrule
\end{tabular}
\end{table}

\begin{figure}[t]
\centering
\includegraphics{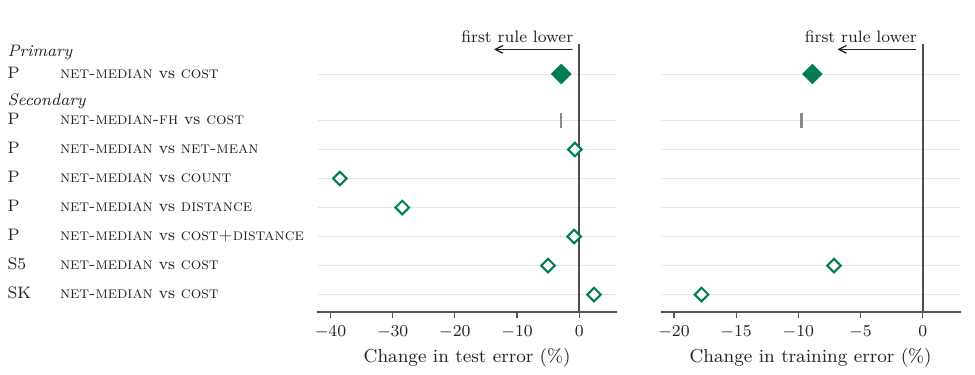}
\caption{Round 2: the pre-registered tests of \cref{tab:round2}, as the estimate of each comparison (geometric-mean
ratio minus one over seeds) for test error (left) and training error (right). Negative: the first rule has the lower
error. Filled: the primary comparison; hollow: secondary; gray tick: \mbox{\OursFH}. P: random starts, 2 epochs, 20 seeds;
S5: 5 epochs, 10 seeds; SK: k-means++ starts, 10 seeds. No intervals are drawn, since the pre-registered analysis
reports estimates, pseudomedians and $p$-values (\cref{tab:round2}).}\label{fig:round2tests}
\end{figure}

\paragraph{Primary.} Both hypotheses were confirmed (\cref{tab:round2,fig:round2tests}). With random starts and two epochs, \Ours{} has
a 2.9\% lower test error and an 8.9\% lower training error than \Cost, on 18 and 20 of the 20 seeds. The test result
rests on Shuttle. Without it, the test error is 0.2\% lower ($p=0.26$, unadjusted), while the training error is still
5.1\% lower (19 of 20 seeds). Per cell, the test error is significantly higher in 7 of the 34 cells, by 0.1 to 0.9\%
(YearPredictionMSD with $k=256$ and $1{,}024$, MNIST and Fashion-MNIST with $1{,}024$, SensIT with both, HIGGS with
$1{,}024$), and significantly lower in 10. The training error is significantly lower in 24 cells and higher in 6, by at
most 0.4\%. \Cref{tab:datasets} and \cref{fig:round2} show the pattern. The gains are large on the datasets with isolated groups and
duplicates (KDD Cup 99's training error, Shuttle), and steady, at 0.2--2.6\%, on the clustered data
(GeoNames, household power, 3D Road, cod-rna, Corel, IJCNN1, Census). The losses are small, at 0.1--0.5\%, on the
diffuse data (YearPredictionMSD, the image datasets, SensIT, HIGGS), where revival matters little.
Indeed, there even \Lib{} is within 1\% of \Cost{} on test error.

\paragraph{Longer Schedules.} With five epochs, the advantage grows. The test error is 5.0\% lower and the training error
7.1\% lower, on all ten seeds, and the test error is 0.9\% lower even without Shuttle. \Mean{} ran away again in this
round. Of its runs, 17 ended with more than a quarter of the centers empty, and its test error on KDD Cup~99 was almost
three times that of \Cost. \Ours{} had no such run at five epochs. Across the whole round, its only run with more than a
quarter of its centers empty was on Shuttle with $k=256$, a dataset on which this happens to every rule.

\paragraph{k-means++ Starts.} With k-means++ starts, \Ours{} has a 2.4\% higher test error than \Cost{} (better on one of
ten seeds) and a 17.8\% lower training error. The difference in test error comes entirely from Shuttle, where the test
error of \Ours{} is 56\% higher, while it is 0.3\% lower on the other 13 datasets. With k-means++ starts on Shuttle,
\Cost{} and \Hyb{} more often keep a center near the isolated test point, whereas the training error of \Ours{} on
Shuttle is 65\% lower.

\paragraph{Against the Other Rules.} \Ours{} has a 38\% lower test error than \Lib{} and a 28\% lower one than \Dist.
Against \Mean{} and \Hyb, the differences in test error with two epochs, 0.7\% and 0.8\% in favor of \Ours, are not
significant. However, \Mean{} is the rule that breaks down with five epochs. \Hyb{} is 7\% worse than \Ours{} on test
error with five epochs and 3\% better with k-means++ starts (descriptive). \OursFH{} is within 0.2\% of \Ours{}
everywhere except the training error with two epochs, where it is 1\% lower. We keep \Ours{} as the default because it
is simpler.

\subsection{Early End of the Move Phase}\label{sec:exp-stop}

\Cref{tab:stop} reports the early stop of \cref{sec:early-stop} on seeds that no other run used (50--59 for two and
five epochs, 50--52 for ten). All four versions of a cell were trained in one process, so that the times are paired.
These runs are exploratory.

The half-life rule costs little and saves a few percent of the running time (\cref{tab:stop} in \cref{app:details}). With two epochs, both
errors rise by 0.06\%, and the running time falls by 1.5\% (median; 2.8\% mean). With five and ten epochs, the errors
rise by about 0.1\% (test) and 0.3\% (training), and the time falls by about 5\%. The rule fires at a median of 21 to
50\% of the move phase, most often on the clustered low-dimensional data (household power, 3D Road, cod-rna, GeoNames), where it saves 8--14\% of the time with five and ten epochs, and almost
never on Shuttle, YearPredictionMSD and SensIT. Most of its cost in training error comes from KDD Cup~99 (+2.3 to +2.9\%
on that dataset). There, it fired in a few runs late in the move phase and missed moves that mattered for the tail of
the training error. The $10^{-4}$ rule saves 3 to 9\% of the time for errors 0.1 to 0.9\% higher. With it, \Ours{} is
about as fast as \Cost{} with five and ten epochs, while keeping most of its advantage (test error 4.2\% and 3.6\%
lower). We recommend the half-life rule as a default. It never changed an error by more than 0.6\% on average on any
dataset except KDD Cup~99, and it removes up to 14\% of the running time on long schedules of low-dimensional data.

\subsection{Round 3: Check Interval and the Freeze}\label{sec:exp-round3}

The third round tested two choices that \Ours{} takes over from \Cost, on all 34 cells with seeds 200--299, which no
earlier run had used (protocol in \cref{app:protocol}). We tested a claim that two settings give the same error by two
one-sided $t$-tests on the mean over seeds of the log ratio, with a margin of $\pm1\%$ fixed before the runs. This margin
is a third of the smallest advantage of \Ours{} over \Cost{} confirmed in round 2. The other tests are those of
\cref{sec:protocol}. Before any new run, 68 runs repeated from round 2 reproduced its errors exactly.

\begin{table}[t]
\centering
\small
\caption{Round 3: pre-registered tests. Estimate: geometric-mean ratio minus one over seeds; pseudomedian in parentheses;
$p$ Holm-adjusted within each family (I: 12 tests; F: 8); seeds: seeds on which the first setting had the lower error.
Same: equivalence within $\pm1\%$ (two one-sided $t$-tests); different: two-sided Wilcoxon. I: driver 98deaca,
2 epochs, 100 seeds. F: driver 79440bc, 2 epochs with 60 seeds and 5 epochs with 40; ``without'' is \Ours{} without the
freeze, ``with'' is \Ours{} with it.}
\label{tab:round3}
\footnotesize
\setlength{\tabcolsep}{4pt}
\begin{tabular}{llllrrl}
\toprule
Comparison & Endpoint & Hypothesis & Estimate & $p$ & Seeds & Result\\
\midrule
\multicolumn{7}{l}{\emph{Family I: check interval of \Ours}}\\
every 10 vs every 5 & test & higher & $+2.4\%$ ($+2.1\%$) & $10^{-13}$ & 13/100 & confirmed\\
every 10 vs every 5 & training & higher & $+2.1\%$ ($+2.1\%$) & $5\cdot10^{-14}$ & 16/100 & confirmed\\
every 20 vs every 5 & test & higher & $+10.4\%$ ($+10.1\%$) & $5\cdot10^{-17}$ & 0/100 & confirmed\\
every 20 vs every 5 & training & higher & $+10.8\%$ ($+10.8\%$) & $5\cdot10^{-17}$ & 0/100 & confirmed\\
\addlinespace[2pt]
every 1 vs every 5 & test & same & $-0.8\%$ ($-0.7\%$) & 0.26 & 65/100 & not shown\\
every 1 vs every 5 & training & same & $+0.2\%$ ($+0.2\%$) & $2\cdot10^{-4}$ & 43/100 & confirmed\\
every 2 vs every 5 & test & same & $-0.4\%$ ($-0.4\%$) & 0.006 & 62/100 & confirmed\\
every 2 vs every 5 & training & same & $-0.7\%$ ($-0.6\%$) & 0.12 & 61/100 & not shown\\
\addlinespace[2pt]
every 1 vs every 5 & test & different & $-0.8\%$ ($-0.7\%$) & 0.002 & 65/100 & lower\\
every 1 vs every 5 & training & different & $+0.2\%$ ($+0.2\%$) & 0.26 & 43/100 & not significant\\
every 2 vs every 5 & test & different & $-0.4\%$ ($-0.4\%$) & 0.08 & 62/100 & not significant\\
every 2 vs every 5 & training & different & $-0.7\%$ ($-0.6\%$) & 0.005 & 61/100 & lower\\
\addlinespace[3pt]
\multicolumn{7}{l}{\emph{Family F: the freeze on the newer driver}}\\
without vs with, 2 epochs & test & lower & $-1.3\%$ ($-1.2\%$) & $6\cdot10^{-9}$ & 56/60 & confirmed\\
without vs with, 2 epochs & training & lower & $-2.3\%$ ($-2.3\%$) & $3\cdot10^{-9}$ & 51/60 & confirmed\\
without vs \Cost, 2 epochs & test & lower & $-4.5\%$ ($-4.6\%$) & $7\cdot10^{-9}$ & 50/60 & confirmed\\
without vs \Cost, 2 epochs & training & lower & $-17.0\%$ ($-17.1\%$) & $10^{-10}$ & 60/60 & confirmed\\
\addlinespace[2pt]
without vs with, 5 epochs & test & same & $+0.1\%$ ($+0.1\%$) & $10^{-31}$ & 9/40 & confirmed\\
without vs with, 5 epochs & training & same & $-0.6\%$ ($-0.5\%$) & $3\cdot10^{-4}$ & 34/40 & confirmed\\
without vs \Cost, 5 epochs & test & lower & $-3.8\%$ ($-3.8\%$) & $7\cdot10^{-8}$ & 35/40 & confirmed\\
without vs \Cost, 5 epochs & training & lower & $-11.2\%$ ($-11.0\%$) & $10^{-11}$ & 40/40 & confirmed\\
\bottomrule
\end{tabular}
\end{table}

\begin{figure}[tb]
\begin{minipage}[c]{3.3in}
\includegraphics{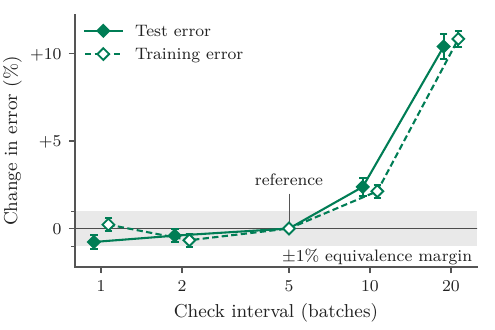}
\end{minipage}\hfill
\begin{minipage}[c]{\dimexpr\linewidth-3.3in-0.25in\relax}
\caption{Round 3, family~I: change in test and training error of \Ours{} when it checks every 1, 2, 10 or 20
batches instead of every 5, as the geometric-mean ratio minus one over 100 seeds (34 cells, 2 epochs; the estimates of
\cref{tab:round3}). Bars: 95\% intervals. Gray band: the equivalence margin of $\pm1\%$.}\label{fig:interval}
\end{minipage}
\end{figure}

\paragraph{Check Interval.} \Ours{} checks every five batches, as \Cost{} does, and nothing in the rule fixes that
number (\cref{sec:limitations}). Checking less often is worse. Every 10 batches raises both errors by more than 2\%, and
every 20 batches by more than 10\% (\cref{tab:round3,fig:interval}). Checking more often helps a little. Every batch lowers the test
error by 0.8\% (95\% interval 0.35 to 1.18\%, so equivalence within 1\% was not shown) and leaves the training error
within 1\%. Every two batches leaves the test error within 1\% and lowers the training error by 0.7\%. These gains come
from KDD Cup~99 and, for the test error, from Shuttle. On the other 12 datasets, checking every 1, 2 or 10 batches
changes neither error by more than 0.25\%, and checking every 20 batches changes them by up to 1.5\%. Denser checks cost
time. With the implementation of round 2, a run took a median 24\% longer when checking every batch and 11\% longer when
checking every two batches, and it took 5\% and 8\% less time when checking every 10 and 20 batches. Five batches is
therefore not where the error is lowest, but a tested compromise. \Cref{app:details} reports exploratory runs of \Cost{}
with other check intervals.

\paragraph{The Freeze on a Newer Driver.}\label{sec:exp-head} The pull request changed after our second round (commit
79440bc). The freeze was replaced by a closing step that reassigns the points of the last pass, and its pair guard was
removed (\cref{app:details} reports exploratory runs on this driver). The third round tested dropping the freeze
on this driver, and all eight of its hypotheses were confirmed (\cref{tab:round3}). Without the freeze, \Ours{} has
a 1.3\% lower test error and a 2.3\% lower training error than with it at two epochs, and both errors within 1\% at five.
At both, it has 3.8 to 17.0\% lower errors than the newer \Cost. Without the freeze, the two-nearest search and the checks continue through the last pass. In
our Python implementation on this driver, a run then took a median 24\% longer with two epochs and 6\% longer with five.

\paragraph{The Released Implementation.}\label{sec:exp-final} We then ran the class that our code release provides for
the current revision, with the C++ parts of \cref{sec:implementation} and the early stop, on the cells, seeds and
machines of part F (\cref{tab:final} in \cref{app:details}; plan in \cref{app:protocol}). It reproduces the research class of the third round
exactly. Without the early stop, all 136 runs at seeds 200--201 equaled their round-3 records, as did all 3{,}400 runs
of \Cost. With the early stop, the results are those of the third round (\cref{tab:final}). In particular, without the freeze
at two epochs, \Ours{} has a 4.4\% lower test error and a 16.9\% lower training error than \Cost. The early stop ended the move phase in about 40\% of the runs at two epochs and in half of them at five, and it
raised the errors by at most 0.3\%. With the C++ parts, dropping the freeze costs little time (\cref{tab:final}), and \Ours{} took a
median 7\% longer than \Cost{} without the freeze at two epochs. In the
research implementation of the third round, \Ours{} without the freeze had taken 50\% and 49\% longer. Our released
implementation therefore drops the freeze by default on this revision, and the freeze remains an option.

\subsection{Exploratory: GPU e-means and o-means}\label{sec:exp-other}

The decision is a function of per-center statistics. It therefore runs unchanged inside the PyTorch implementation of
e-means in the pull request and inside o-means~\citep{bajaj2026omeans}, which keeps the count rule of e-means and adds a
two-level assignment. Our optimized versions run the whole check on the GPU (\cref{app:details}). Given the same state,
they made the same decisions as the C++ decision in every
check we compared (more than 5{,}000). On the current revision of the driver, whole runs reached the same final centers,
bit for bit, as an independent proof of concept whenever both used the same nearest-center search. In o-means, the
second-nearest distance comes from o-means' own two-level search, so it is approximate where that search is. All runs
used one RTX 3090 Ti, and none is pre-registered (\cref{tab:gpu} in \cref{app:details}).

\paragraph{Accuracy.} On the earlier revision of the driver, the GPU version lowers the training error of the GPU rule of
the pull request by 11.6\% (on all ten seeds) and leaves its test error unchanged ($+0.4\%$, not significant). On the current revision, dropping the freeze lowers the test and
training errors at two epochs, by 1.7\% and 4.2\%, and changes neither by more than 0.2\% at five epochs, as on the CPU
(\cref{sec:exp-head}). On DEEP and SPACEV with one million points and $k=16{,}384$, the rule without the freeze is level
with \Cost, while with the freeze it is slightly worse on all five seeds. With the freeze, about 300 of the 16{,}384
centers end the run empty, against 40 without it and 18 with \Cost. On ten million points (DEEP, three seeds), the three
rules are within 0.06\% of each other on both errors, and each leaves 11 to 16 centers empty. Inside o-means, the rule
lowers the test error of o-means' own rule by a third and the training error by 40\%.

\paragraph{Cost.} These runs last 20 to 400 milliseconds, and the rule adds 10 to 170 milliseconds to each. This makes
them 33 to 66\% longer than with the GPU rule of the pull request, and 73\% longer than with the rule of o-means
(\cref{tab:gpu}). Without the freeze, the checks continue through the last pass, about twice as many at two epochs,
hence the higher ratio. On ten million points (DEEP, $k=16{,}384$, the GPU not shared with other jobs), the rule added
7\% to the running time with the freeze and 12\% without it. Inside o-means, runs with the rule take 3.2 times as long as with o-means' own rule on one
million points and 2.9 times as long on ten million, where o-means' own assignment is fast. \Cref{app:details} explains why the overhead falls with the data in e-means but
not in o-means.

\section{Related Work}\label{sec:related}

We review two lines of work, namely the relocation of centers in clustering and vector quantization, and decisions
based on noisy estimates.

\paragraph{Relocating Centers.} Moving the centers of useless clusters to where they help is an old idea in vector
quantization and clustering. Fritzke's utility, the increase in error if a center were removed, drives the relocations
of LBG-U~\citep{fritzke1997lbgu}, of the growing neural gas with utility~\citep{fritzke1997gngu} and of breathing
k-means~\citep{fritzke2020breathing}. ELBG~\citep{patane2001elbg} moves low-utility centers next to high-distortion ones,
and random swap~\citep{franti2018randomswap} tries relocations and keeps those that lower the error. Mini-batch k-means
in scikit-learn reassigns centers with small cumulative counts~\citep{sculley2010,sklearn_minibatch_code}, and the
codebooks of vector-quantized neural networks are revived by similar count rules or by
re-initialization~\citep{dhariwal2020jukebox,zeghidour2021soundstream,lancucki2020robust,zheng2023cvq}. In
e-means~\citep{bessa2026emeans}, revival went from counts (\Lib) to counts with a distance test
(\Dist;~\citep{weiss2026small}), a separation test~\citep{faiss2026pr5908}, and estimated cost and gain
(\Cost;~\citep{faiss2026pr5915}), while o-means~\citep{bajaj2026omeans} keeps the count rule.
Assylbekov~\citep{assylbekov2026counterexamples} gives two counterexamples to e-means, one of them about revival, which
the distance test fixes and \Cost{} does not. The closest line of work is \Cost, which estimates the same two quantities
as our rule but compares the raw estimates. All these methods compare raw statistics or try moves on the data, and none
of them accounts for how much evidence each statistic rests on, in contrast to the rule reported here.

\paragraph{Decisions on Noisy Estimates.} Selecting the largest of many noisy estimates overstates the selected value,
an effect known as the optimizer's curse~\citep{smith2006optimizer}. The remedy that this work recommends is to replace
each estimate by its posterior given a prior over the population of estimates, which is what our rule does. Shrinking
unit-level estimates toward a population fitted to the data is parametric empirical Bayes~\citep{morris1983parametric}.
Our gain model is the random-effects model of meta-analysis. Its between-unit variance is estimated by the method of
moments~\citep{dersimonian1986meta} or, as recommended when the precisions of the units vary
widely~\citep{veroniki2016methods}, by the method of Paule and Mandel~\citep{paule1982consensus}. The covariate version
is the Fay--Herriot model of small-area estimation~\citep{fay1979estimates}. That a posterior mean of a log-normal
quantity is dominated by its spread when evidence is absent is elementary. That this effect decides revival in practice
has not been reported before.

\section{Limitations}\label{sec:limitations}

We list the limitations of the rule and of our evaluation.

\paragraph{The Value of a Move Is a Surrogate.} The net gain ignores the interactions between a removal and a split, and
between the moves of a check. It also ignores how the centers adapt afterwards. The matching is optimal for the
surrogate only under the conditions of \cref{prop:greedy}. We do not know how far the surrogate is from the real effect
of a move in general. Immediate measurements understate that effect where centers adapt strongly (the technical
report~\citep{weiss2026report} gives an example).

\paragraph{The Models Are Working Models.} The counts are not Poisson, the batches are not independent across epochs,
the gains are Gaussian only approximately, the prior is fitted from the same data it is applied to, and the variances are
plugged in as known. The calibration (\cref{app:calibration}) shows that the selected gains are still overestimated,
although by less than the raw estimates. That calibration used the posterior means of the first version, and the median
version was not recalibrated. Every correction of the remaining bias that we tried made the rule worse.

\paragraph{The Check Interval Is Not Derived.} The rule weighs the evidence for each move within a check, but nothing in
it says how often to check. \Ours{} keeps the interval of \Cost, five batches, and the arguments for any interval are
soft. Frequent checks cost time and move centers more often, which can make training less stable, while rare checks
leave fewer chances to correct a poor placement before the step size has shrunk. The third round
(\cref{sec:exp-round3}) supports five batches on our suite, but it does not show that five batches is right for other
batch sizes, schedules or data. An interval set by the
evidence itself is left for future work.

\paragraph{Test Error with Isolated Test Points.} On Shuttle, and with k-means++ starts in particular, \Ours{} keeps
fewer centers near the one test point far from all training data, and its test error there is much higher than that of
\Cost{} and of \Hyb. Training data cannot reveal such a point. A rule that keeps more centers in sparse regions (e.g.,
the distance test) does better on it by design and worse on the training error. On test error, \Ours{} is not better
than \Hyb{} with two epochs.

\paragraph{Embeddings at Large k.} On DEEP and SPACEV with $k=16{,}384$, the rule is within a few tenths of a percent of
\Cost, as revival matters little there. With the freeze, it leaves more centers empty than \Cost{} and is slightly
worse. Without the freeze, it is level with \Cost{} (\cref{sec:exp-other}).

\paragraph{Scope.} All three rounds used one family of algorithms (e-means and the drivers in the pull request) on
CPUs. The early stop and the GPU and o-means versions are exploratory, and the final comparison of the released
implementation is descriptive. The exact equivalence of the C++ decision is tied to the versions of NumPy and OpenBLAS
that we used.

\paragraph{Very Large Problems.} We ran the GPU versions on one consumer GPU, with up to ten million points,
$k\le16{,}384$ and exact assignment. The largest runs of e-means~\citep{bessa2026emeans} use several H100 GPUs, up to ten
billion points and 16.7 million centers, approximate (IVF) assignment and 8-bit inputs. Porting the rule there is
engineering that we have not done (\cref{app:details} lists its three parts). Two questions are open beyond the engineering. The first is whether the
decisions stay good when the margins come from an approximate assignment. The second is whether revival pays at that
scale at all, since at the largest scale we measured the rule was level with \Cost.

\section{Conclusions}\label{sec:conclusion}

We considered revival in mini-batch k-means, whose decisions compare estimates that differ in quality by orders of
magnitude between centers. Treating these estimates alike lets the least informed centers decide. Raw comparisons act
on noise, a posterior mean rewards ignorance, and a moment prior is set by the noisiest units. We showed that weighting
every estimate by its evidence and deciding by posterior medians gives a rule without tuning parameters, \Ours. In a
pre-registered comparison, it lowered both the test and the training error of \Cost, the strongest earlier rule on our
benchmarks. It stayed stable on longer schedules, where its predecessor broke down, and it cost a few percent of running
time. The rule is a function of per-center statistics only, so it can be used unchanged in other implementations of
mini-batch k-means. How often to check remains open, and so does whether revival pays at the largest scales of e-means.
It would be interesting to set the check interval by the evidence itself, and to port the rule to multi-GPU e-means with
approximate assignment.

\section*{AI Use Statement}

Claude Opus~5.5 (Anthropic), used through Claude Code, was used for coding, experiments and writing. GPT-6.1-sol
(OpenAI), used through Codex, was used for reviews of the mathematics, the code and the experimental protocols. The
authors reviewed all AI-assisted text, figures, and code. We take responsibility for the final content of this paper,
including all claims.

\bibliographystyle{plainnat}
\bibliography{refs}

\appendix
\crefalias{section}{appendix}
\section{Complete Specification of the Rule}\label{app:spec}

This appendix states \Ours{} exactly as implemented, including every floor and boundary case. The reference
implementation is in \texttt{ablation.py} and \texttt{ngp\_variants.py}, and the C++ decision in
\texttt{ngp\_decide.cpp}.

\paragraph{Statistics.} For each center, the rule keeps the following statistics since its birth: $a_j$ batches, $n_j$
points, and, over the points received while moves are allowed, $M_j=\sum m$, $Q_j=\sum m^2$, $P^{(2)}_j=\sum p^2$ and
$P^{(4)}_j=\sum p^4$, where $p$ is the projection on the center's split axis. With the two-nearest search on every point
while moves are allowed, $M_j$ and $Q_j$ cover the same $n_j$ points. The rule also keeps the inherited coefficient
$h_j$, which is multiplied by $1-\alpha_t$ after every batch. Globally, it keeps $\Pi_t=\prod_{u\le t}(1-\alpha_u)$ since
the start of training, with $t$ the zero-based index of the batch. A move $(j,s)$ sets $a$, $n$, $M$, $Q$, $P^{(2)}$ and
$P^{(4)}$ of both $j$ and $s$ to zero and their $h$ to 1. Their usage and split statistics are those that the split of
\Cost{} gives them (the donor's, halved), and $\Pi_t$ is not changed. Throughout, $\mathrm{mean}$ and $\mathrm{var}$ are
over the stated set, and $\mathrm{var}$ divides by the number of its elements.

\paragraph{Split.} As in \Cost, the two halves are $c_s\pm a_su_s$, with $u_s=U_s/\lVert U_s\rVert$ and
$a_s=\sqrt{2\lVert U_s\rVert/(\pi N_s)}$, when $\lVert U_s\rVert$ and $N_s$ are finite and positive. Otherwise, the
default separation of the two copies in \Cost{} is used. The rule changes only which moves are made.

\paragraph{Rates.} Let $S=\{j:a_j\ge1\}$. If $|S|<2$, then $\tilde\lambda_j=n_j/a_j$ for $a_j>0$, and 0 for the other
centers. Otherwise, let $\rho_j=n_j/a_j$ on $S$ and $\bar\lambda=\mathrm{mean}(\rho)$. If $\bar\lambda\le0$, every rate
is 0. Else, $v=\max\{\mathrm{var}(\rho)-\mathrm{mean}(\rho_j/a_j),\,10^{-6}\bar\lambda^2\}$, $a=\bar\lambda^2/v$,
$b=\bar\lambda/v$, and $\tilde\lambda_j=(a+n_j)/(b+a_j)$ for every $j$.

\paragraph{Distances.} The rule takes $\delta_j$ and $n(j)$ from an exact two-nearest search of the centers among
themselves (FAISS). If no other center is found or the distance is not finite, $\delta_j$ is set to the finite value
$D_{\max}=1.8\cdot10^{308}/4$ (i.e., a quarter of the largest double), and if no other center is found, $n(j)=j$.

\paragraph{Margin Ratios.} Let $O=\{j:n_j>0,\ \delta_j>0,\ \delta_j\text{ finite}\}$ (a center with
$\delta_j=D_{\max}$ belongs to $O$). If $O$ is empty, $\tilde r_j=1$ for every $j$. Otherwise, $\hat r_j=M_j/(n_j\delta_j)$
on $O$ and $\bar r=\sum_On_j\hat r_j/\sum_On_j$. If $|O|=1$, then $\tilde r_j=\max(\bar r,0)$ for every $j$. Else, over
$O_2=\{j\in O:n_j\ge2\}$, $\sigma^2=\max\{\sum_{O_2}(Q_j-M_j^2/n_j)/\delta_j^2\,/\sum_{O_2}(n_j-1),\ 10^{-12}\}$ (if $O_2$
is empty, $\sigma^2=\max\{\mathrm{var}_O(\hat r),10^{-12}\}$), and $\tau_r^2=\max\{\mathrm{var}_O(\hat r)-\mathrm{mean}_O(\sigma^2/n_j),\
10^{-6}\bar r^2,\ 10^{-12}\}$. The ratio $\tilde r_j$ is given by \cref{eq:rhat} on $O$ and by $\bar r$ elsewhere, and it
is then clipped below at 0. The cost is $\tilde C_j=\tilde\lambda_j\tilde r_j\max(\delta_j,0)$.

\paragraph{Gains.} The raw gain is $\hat G_s=\tfrac2\pi\lVert U_s\rVert/\max(1-\Pi_t,10^{-12})$. The kurtosis is pooled
over $K=\{j:n_j\ge2,\ P^{(2)}_j>0\}$, with $\kappa_j=(P^{(4)}_j/n_j)/(P^{(2)}_j/n_j)^2$ and
$\bar\kappa=\sum_Kn_j\kappa_j/\sum_Kn_j$, and $\bar\kappa-1$ is floored at $10^{-6}$ (and set to 2 if $K$ is empty).
With $\alpha_t$ clipped to $[10^{-6},1]$, $W_t=\min\{(2-\alpha_t)/\alpha_t,\ t+1\}$ and
$v_s=(\bar\kappa-1)/\max(\tilde\lambda_sW_t,10^{-3})$. Let $P=\{s:\hat G_s>0\}$ and $m=|P|$. A center outside $P$ has
gain 0. If $m<2$, then $\tilde G_s=\hat G_s$. Otherwise, the rule sets $y_s=\log\hat G_s$ on $P$ and fits the
Paule--Mandel prior. With $w_s(t)=1/(t+v_s)$, $\mu(t)=\sum w_sy_s/\sum w_s$ and $Q(t)=\sum w_s(y_s-\mu(t))^2$, if
$Q(0)\le m-1$, then $\tau^2=0$. Otherwise, $\tau^2$ approximates the root of $Q(\tau^2)=m-1$ by bisection on $\log t$
from $[-30,\ \log\max(4\,\mathrm{var}(y),10^{-12})+1]$, with the upper end raised by 2 until $Q\le m-1$ or it reaches
60, and with 100 halvings. Then $\tau^2=\exp$ of the midpoint of the final bracket (a root below $e^{-30}$ is returned as
about $e^{-30}$), and $\mu=\mu(\tau^2)$ at that value. For the posterior only, $\tau^2=\max(\tau^2,10^{-6})$, with $\mu$
unchanged. Finally, $\tilde G_s=\exp\bigl((\tau^2y_s+v_s\mu)/(\tau^2+v_s)\bigr)$.

\paragraph{Matching.} The donors are the centers with $N_s>0$ and $h_s\le\frac12$, sorted by $\tilde G_s$ decreasing
(ties by index). The candidates are the centers with $h_j\le\frac12$ and $\tilde C_j<\tilde G$ of the first donor,
sorted by $\tilde C_j$ increasing (ties by index). The rule visits the candidates in order. For candidate $j$, let $s$
be the first donor that is neither freed nor used nor $j$. If there is none, the rule stops if every remaining donor is
freed or used, and otherwise goes to the next candidate. If $\tilde C_j\ge\tilde G_s$, it stops if $s$ is the first
available donor, and otherwise goes to the next candidate (the better donor was $j$ itself). If $j$ is protected, or if
$n(j)$ was freed at this check and $c_j\ne c_{n(j)}$, or if $j$ was used as a donor, it goes to the next candidate.
Otherwise, it frees $j$, protects $n(j)$, uses $s$, and records the move $(j,s)$.

\paragraph{When the Guard Makes the Greedy Matching Suboptimal.} Consider five centers with costs
$(1,1.1,1.2,100,100)$, gains $(0,0,0,10,9)$, $n(1)=2$ and $n(2)=n(3)=1$. The greedy matching frees center 1 for donor 4
(value 9) and must then skip centers 2 and 3, whose nearest center was freed. However, freeing centers 2 and 3 for
donors 4 and 5 is allowed and has value 16.7.

\paragraph{The Fay--Herriot Variant (\OursFH).} Only the prior mean of the log gain changes. The model is
$y_s=x_s^\top\beta+\eta_s+e_s$, with $\eta_s\sim\mathcal N(0,\tau^2)$, $e_s\sim\mathcal N(0,v_s)$ and the covariates
$x_s=(1,\log\max(\tilde\lambda_s,10^{-6}),\log\max(\delta^-_s,10^{-12}))$, where $\delta^-_s$ is the center's $\delta$
at the previous check (the first check uses the first two covariates only). For given $t$, $\beta(t)$ is the
least-squares solution of $(X^\top WX+10^{-12}I)\beta=X^\top Wy$ with $W=\mathrm{diag}(w_s(t))$. Let
$Q(t)=\sum w_s(y_s-x_s^\top\beta(t))^2$, and let $p$ be the number of covariates. If $Q(0)\le m-p$, then $\tau^2=0$ and
$\beta=\beta(0)$. Otherwise, $\tau^2$ approximates the root of $Q(\tau^2)=m-p$ by the same bisection with 60 halvings,
and $\beta=\beta(\tau^2)$. For the posterior only, $\tau^2=\max(\tau^2,10^{-6})$, with $\beta$ unchanged, and
$\tilde G_s=\exp\bigl((\tau^2y_s+v_sx_s^\top\beta)/(\tau^2+v_s)\bigr)$. With $m<4$, the raw gains are used.

\section{Calibration of the Estimates}\label{app:calibration}

During development, we compared the predicted cost and gain of every move and of 64 random centers with their exact
first-order values on the full training and test sets. We did so at every check of 21 runs (six datasets with
$k=1{,}024$ and Shuttle with $k=256$, three seeds), for 51{,}569 comparisons in all. Three findings shaped the rule.

\emph{Low-evidence costs are strongly underestimated by raw statistics.} For centers whose running count was below
a tenth of the memory of the EMA, the bound $N_j\delta_j$ (a cost estimate from the running count) was a median 13 times
below the exact removal cost. For the centers that were freed, it was 100 times below. Such moves were 27--57\% of
all moves on KDD Cup~99, YearPredictionMSD and Shuttle. With posterior rates and shrunk margin ratios, the cost of
random centers became unbiased to within 0.02--0.11 in $\log_{10}$, and that of freed centers was 0.08--0.18 too low.

\emph{The raw gain is calibrated for random centers and biased for the chosen ones.} For centers with enough
evidence, the predicted gain matched the exact gain (median $\log_{10}$ ratio $-0.03$ to $+0.08$). However, the gains of
the chosen donors were 1.2 to 2.1 times too high, an instance of the optimizer's curse. Shrinkage reduced this bias (for
the chosen donors with the least evidence, from $+0.23$ to $0.00$ in $\log_{10}$) but did not remove it (up to 1.9 times
on YearPredictionMSD).

\emph{Training and test agree once evidence exists.} The exact values on training and test data differed by a
median of 0 once a center had at least about one point in the memory of the EMA. They diverge only for the low-evidence
centers.

These calibrations used the posterior means of the first version, and the median version was not recalibrated.

\section{The Decision in C++}\label{app:cpp}

\paragraph{Comparison on All Cells.} Our reference implementation computes the estimates of \cref{sec:estimation} with
NumPy. For the confirmatory study, the per-point sums and the matching were in C++, and the per-check estimates stayed in
Python. We compared the two on the 34 dataset--size combinations of \cref{sec:experiments}, with two and five epochs and
three versions of the rule, 17{,}300 checks in all. On a processor without AVX-512, every estimate was bit-identical in
all 7{,}700 checks made there.

\paragraph{Implementation and Tests.} The C++ decision (\texttt{ngp\_decide.cpp}, with a Python subclass \texttt{ngp\_fast.py}) receives the per-center
statistics and the two-nearest search of the centers by FAISS, and it returns the moves. It replicates the pairwise
summation of NumPy~2.5 (eight accumulators, blocks of 128, halving split), the order of every arithmetic operation, and
the tie rules of Python's \texttt{max} and NumPy's \texttt{maximum}. It is compiled without fused multiply-adds, and it
calls the same OpenBLAS and LAPACK routines as NumPy for the least-squares problems of \OursFH. Exponentials and
logarithms come from the C library. Each of 306 test jobs trained one configuration twice in one process, once with the
reference and once with the C++ decision. Each job compared the rates, distances, costs, gains, population ratio mean
$\bar r$ and number of candidates bit by bit at every check, as well as the moves and final errors. All 306 jobs made
identical moves at every check and reached identical final errors. On the machine without AVX-512, every estimate was
bit-identical. On the machine with AVX-512, only the gains differed, by at most $7\cdot10^{-15}$ relative
($5\cdot10^{-13}$ for \OursFH), because NumPy uses its own vectorized exponential and logarithm there. The decision took
0.6 to 2.2\,ms per check in the reference and 0.12 to 0.63\,ms in C++ (\Ours). Its share of the running time fell
from a median of 2.2\% to 0.6\% (from 31\% to 5\% on the smallest dataset). An independent review in two rounds found no
remaining defect of consequence.

\section{Protocols}\label{app:protocol}

\paragraph{Equivalence Gates of the C++ Version.} In each
round, an equivalence part compared it with the pure Python reference on separate seeds, and the analysis was allowed to
run only if every result agreed to a relative $10^{-9}$. In both rounds, all results were identical. The protocols, the
analysis code, the code identities and every run record are in the repository.

\paragraph{Round 1.} The protocol was written before the runs. \Mean{} was frozen, and its C++ parts were checked
against the reference (24 whole runs bit-identical, 3{,}000 matchings identical). The primary test was on test error,
with random starts, two epochs, 16 cells and seeds 10--29, by a two-sided Wilcoxon test at 5\% with the estimate below
zero. The secondary tests formed one Holm family of eight. They covered the comparisons with \Lib, \Dist{} and \Hyb,
k-means++ starts (seeds 0--9), Covertype with random and k-means++ starts, and DEEP, SIFT and SPACEV with $k=16{,}384$
against \Cost{} and \Lib. Five epochs, sampling ablations and timing were descriptive. There were two amendments. A
part on DEEP~10M was stopped for lack of machine time, and a stale lock was removed.

\paragraph{Round 2.} The protocol was written before the runs and after the candidate was chosen, with \Ours{} as the
primary rule and \OursFH{} as the alternative. It had the parts P (34 cells, seeds 30--49, random starts, 2 epochs;
4{,}760 runs), S5 (5 epochs, seeds 30--39) and SK (k-means++ starts, seeds 30--39), and two equivalence parts against
the reference (all identical). In all, it made 9{,}520 runs, with no error. The two co-primary endpoints were corrected by
Holm over two, and the secondary tests formed one Holm family of ten (\cref{tab:round2}). Adverse cells were identified
by per-cell Wilcoxon tests with Holm over 34 cells, and the direction rule of \cref{sec:protocol} applied. Every run
record carries the hashes of the code, the data and the start, and the analysis refuses mixed records. Datasets were
pinned to machines because KDD Cup~99 gives slightly different results on the two processors (each machine is
deterministic). There was no amendment.

\paragraph{Round 3.} The protocol was written before the runs, reviewed independently, and sized by a power analysis on
the exploratory runs, which raised the planned 20 seeds to 100, 60 and 40. The round used seeds 200--299, all 34 cells,
random starts and the machines of round 2. As a gate, 68 runs repeated from round 2 (seeds 30--31) had to reproduce its
errors exactly before any new run, and all were identical. Part I ran \Ours{} checking every 1, 2, 5, 10 and 20 batches
on driver 98deaca, with two epochs and 100 seeds (17{,}000 runs). Its tests formed one Holm family of 12, namely that
checking every 10 and 20 batches gives higher errors than every 5, and that checking every 1 and 2 batches is equivalent
to every 5 within $\pm1\%$ and different from it. Part F ran \Ours{} with and without the freeze and \Cost{} on driver
79440bc, with two epochs and 60 seeds and with five epochs and 40 seeds (10{,}200 runs). Its tests formed one Holm family
of 8. At two epochs, the hypotheses were that \Ours{} without the freeze has lower errors than with it and than \Cost,
and at five epochs, that it is equivalent to \Ours{} with the freeze and has lower errors than \Cost. Equivalence was tested
by two one-sided $t$-tests on the mean log ratio. A simulation on the exploratory runs put their error rate at the
margin at 0.5 to 1.8 times the nominal level. Every equivalence shown has a Holm-adjusted $p$ below 0.025, where that
inflation does not change the decision. Every run record carries the hashes of the code and of both data files, and the
analysis refuses mixed or incomplete records. One machine hung during part I. After its reboot, every record was
checked, and the missing runs were resumed with unchanged code. There was no amendment, and the round made 27{,}268
runs with no error.

\paragraph{Final Comparison.} This was not a confirmatory round. Its analysis plan was written after the launch and
before any result was read, and it estimates effects without tests of hypotheses. The released class for the current
revision of the driver, with its C++ parts, ran on the 34 cells and on the seeds, starts and machines of part F of round
3. It ran \Cost{} and \Ours{} with the early stop, with and without the freeze, at two epochs (seeds 200--259) and five
(200--239). Before it, the released class reproduced the round-3 research class bit for bit in all 256 comparisons on
the 14 datasets. It did so also with the early stop, against a reference written separately. \Ours{} without the early stop
(seeds 200--201) and every run of \Cost{} reproduced their round-3 records exactly. The comparison made 10{,}336 runs
with no error. One change was made after the start and recorded in the plan before the five-epoch runs ended. The
running time of the comparison with the runs of the third round is not reported, because those runs used a slower
implementation.

\section{Additional Details of the Experiments}\label{app:details}

This appendix holds details that support \cref{sec:experiments,sec:limitations}.

\paragraph{Round 1 by Dataset (Test Error Against \Cost).} \Mean{} was better on Census, GeoNames and household power by 0.6
to 2.5\% and on Shuttle with $k=256$ by 35\%. However, it was worse on KDD Cup~99 with $k=256$ and $k=4{,}096$, by 25\%
and 15\%, and slightly worse on YearPredictionMSD.

\paragraph{Check Interval of \Cost{} (Exploratory).} In exploratory runs (seeds 0--4), \Cost{} was worse both when it
checked every batch (training error 12\% higher than at five) and when it checked every 20 batches. The advantage of
\Ours{} over \Cost{} grew as checks became more frequent (training error 22.5\% lower when both checked every batch). An
exact test that skips the checks that cannot accept a move turned out not to skip any while moves were allowed.

\paragraph{The Newer Driver (Exploratory).} Revision 79440bc of the pull request replaced the freeze by a closing step
and removed the pair guard (\cref{sec:exp-head}). In exploratory runs (seeds 30--49), this raised the training error of \Cost{} by 6.7\% (on KDD Cup~99 with
$k=1{,}024$ and $4{,}096$, by 79\% and by a factor of 4.01). On this driver, \Ours{} with its freeze and guard
reproduces the round-2 runs, and dropping our pair guard raised its errors.

\paragraph{GPU Implementation.} Our first, unoptimized versions computed the statistics with plain tensor operations and the
decision on the host. On the 14 timing cells of \cref{tab:gpu}, they took 7.5 times as long as the GPU rule of the pull
request and 9 times as long as o-means' own rule. The optimized versions compute the per-point sums inside the captured
CUDA graphs of the drivers and run the whole check on the GPU. This covers the search for every center's nearest other
center (an FP32 screen with a rigorous error bound, then FP64 among the kept candidates), the estimates in fused kernels,
the Paule--Mandel fit and the matching.

\paragraph{GPU Cost and Data Size.} The extra work has two parts that scale differently. The per-point sums grow with the data, like
the driver's own work. The checks cost a fixed time each (about 10 milliseconds at $k=16{,}384$, mostly the
nearest-other-center search), and e-means makes few of them on large data, because its batch grows almost in proportion
to the data. With the freeze, it makes 6 checks on one million points and 7 on ten million, and without it 12 and 15.
On larger data, the overhead should approach the share of the per-point sums, a
few percent. In contrast, o-means checks every five batches of a fixed size, so its number of checks grows with the data
and the overhead does not fall.

\paragraph{Porting to Very Large Problems.} First, the per-center sums must be added across GPUs. Second, the
drivers for that scale do not compute the second-nearest distances that the margins need (we added them to the kernels
of o-means in the same way). Third, the exact search of a check for every center's nearest other center compares all
pairs of centers ($O(k^2d)$). It takes seconds with a million centers in 96 dimensions, but it would take of the order
of an hour per check with 16.7 million centers in 1{,}024 dimensions. There, a search over the index of the centers that
IVF assignment already builds should replace it.

\begin{table}[t]
\centering
\caption{Early end of the move phase (exploratory). Change in test / training error relative to \Ours{} without the
early stop (geometric means over seeds and cells), median paired running time, and the share of runs in which the
stop fired. Half-life: quiet = no move; $10^{-4}$: quiet = net gain below $10^{-4}$ of the batch error. The last column
compares the running time of \Ours{} with that of \Cost.}\label{tab:stop}
\footnotesize
\setlength{\tabcolsep}{4pt}
\begin{tabular}{lcccccccc}
\toprule
 & \multicolumn{3}{c}{Half-life (no move)} & & \multicolumn{3}{c}{Net gain $<10^{-4}$ of the error} & \Ours{} vs \Cost\\
\cmidrule{2-4}\cmidrule{6-8}
Epochs (seeds) & Test / training & Time & Fired & & Test / training & Time & Fired & Time\\
\midrule
2 (10) & +0.06 / +0.06\% & $-$1.5\% & 38\% & & +0.08 / +0.07\% & $-$3.1\% & 46\% & +6.0\%\\
5 (10) & +0.07 / +0.33\% & $-$4.6\% & 50\% & & +0.40 / +0.91\% & $-$9.4\% & 66\% & +7.5\%\\
10 (3) & +0.12 / +0.31\% & $-$4.9\% & 62\% & & +0.24 / +0.63\% & $-$7.6\% & 74\% & +5.7\%\\
\bottomrule
\end{tabular}

\end{table}

\begin{table}[t]
\centering
\small
\caption{Final comparison of the released implementation (driver 79440bc, C++ parts, early stop on; 34 cells; the
seeds of part F). Estimate: geometric-mean ratio minus one over seeds with a 95\% $t$ interval; seeds: seeds on which the
first setting had the lower error; time: median of the paired ratios of running time. ``Early stop vs none'' pairs each
run with the third round's run of the research class without the early stop (same cell and seed); its time is not
comparable (another implementation). Descriptive: no hypothesis is tested.}
\label{tab:final}
\footnotesize
\setlength{\tabcolsep}{4pt}
\begin{tabular}{lllllr}
\toprule
Comparison & Test error & Seeds & Training error & Seeds & Time\\
\midrule
\multicolumn{6}{l}{\emph{2 epochs, 60 seeds}}\\
no freeze vs freeze & $-1.24\%$ $[-1.53, -0.94]$ & 56/60 & $-2.26\%$ $[-2.73, -1.78]$ & 51/60 & 1.06\\
no freeze vs \Cost & $-4.40\%$ $[-5.30, -3.49]$ & 50/60 & $-16.93\%$ $[-17.66, -16.19]$ & 60/60 & 1.07\\
freeze vs \Cost & $-3.20\%$ $[-4.16, -2.23]$ & 47/60 & $-15.01\%$ $[-15.85, -14.16]$ & 60/60 & 1.02\\
no freeze: early stop vs none & $+0.06\%$ $[+0.05, +0.08]$ & 4/60 & $+0.06\%$ $[+0.04, +0.08]$ & 1/60 & ---\\
freeze: early stop vs none & $+0.05\%$ $[+0.03, +0.06]$ & 6/60 & $+0.05\%$ $[+0.05, +0.06]$ & 0/60 & ---\\
\addlinespace[3pt]
\multicolumn{6}{l}{\emph{5 epochs, 40 seeds}}\\
no freeze vs freeze & $+0.10\%$ $[+0.05, +0.14]$ & 11/40 & $-0.45\%$ $[-0.65, -0.26]$ & 35/40 & 1.01\\
no freeze vs \Cost & $-3.77\%$ $[-4.84, -2.69]$ & 35/40 & $-10.96\%$ $[-12.07, -9.84]$ & 40/40 & 1.04\\
freeze vs \Cost & $-3.86\%$ $[-4.93, -2.78]$ & 35/40 & $-10.55\%$ $[-11.64, -9.46]$ & 40/40 & 1.04\\
no freeze: early stop vs none & $+0.08\%$ $[+0.05, +0.10]$ & 4/40 & $+0.30\%$ $[+0.13, +0.48]$ & 1/40 & ---\\
freeze: early stop vs none & $+0.08\%$ $[+0.05, +0.11]$ & 3/40 & $+0.21\%$ $[+0.04, +0.37]$ & 2/40 & ---\\
\bottomrule
\end{tabular}
\end{table}

\begin{table}[t]
\centering
\small
\caption{Exploratory GPU runs (one RTX 3090 Ti). Data: the 34 cells, or DEEP and SPACEV with one million points and
$k=16{,}384$ (1M). Errors: geometric-mean ratio minus one over seeds; lower: seeds on which the first rule had the lower
test and training error. Time: the first rule's time over the second's at two epochs (minimum of three fits per seed,
median over seeds, geometric mean over 14 cells, or over the two 1M cells). GPU e-means is \Ours{} in the pull request's
PyTorch driver; ``no freeze'' and ``freeze'' are \Ours{} on its current revision. Dashes mark comparisons that were not
run.}
\label{tab:gpu}
\footnotesize
\setlength{\tabcolsep}{3.5pt}
\begin{tabular}{lcccrrcr}
\toprule
Comparison & Data & $T$ & Seeds & Test & Training & Lower & Time\\
\midrule
GPU e-means vs GPU \Cost{} (98deaca) & 34 & 2 & 10 & $+0.4\%$ & $-11.6\%$ & 5/10, 10/10 & 1.38\\
no freeze vs freeze & 34 & 2 & 5 & $-1.7\%$ & $-4.2\%$ & 5/5, 5/5 & 1.25\\
no freeze vs freeze & 34 & 5 & 3 & $+0.1\%$ & $-0.2\%$ & 1/3, 2/3 & ---\\
no freeze vs GPU \Cost & 34 & 2 & --- & --- & --- & --- & 1.66\\
no freeze vs GPU \Cost & 1M & 2 & 5 & $-0.08\%$ & $+0.01\%$ & 4/5, 0/5 & 1.67\\
freeze vs GPU \Cost & 34 & 2 & --- & --- & --- & --- & 1.33\\
freeze vs GPU \Cost & 1M & 2 & 5 & $+0.15\%$ & $+0.18\%$ & 0/5, 0/5 & 1.37\\
\addlinespace[3pt]
o-means: \Ours{} vs its own rule & 34 & 2 & 10 & $-32.4\%$ & $-40.3\%$ & 10/10, 10/10 & 1.73\\
o-means: \Ours{} vs its own rule & 1M & 2 & 5 & $-0.4\%$ & $-0.5\%$ & 5/5, 5/5 & 3.17\\
\bottomrule
\end{tabular}
\end{table}

\end{document}